\documentclass[runningheads]{llncs}
\usepackage{graphicx}
\usepackage{amsmath,amssymb} % define this before the line numbering.
\usepackage{color}
\usepackage{xcolor}
\usepackage{graphicx}
\usepackage{subcaption}
\usepackage{float}
\graphicspath{ {./images/} }
\usepackage[width=122mm,left=12mm,paperwidth=146mm,height=193mm,top=12mm,paperheight=217mm]{geometry}
\usepackage{hyperref}

\begin{document}
% \linenumbers
\pagestyle{headings}
\mainmatter

\title{2D Image Relighting with Image-to-Image Translation} % Replace with your title

\institute{EPFL, Switzerland \footnote{Project supervised by Majed El Helou in CS413 at EPFL.}\\\href{mailto:paul.gafton@epfl.ch}{\email{\{paul.gafton,}}\href{mailto:erick.marazzuniga@epfl.ch}{\email{erick.marazzuniga\}}}\email{@epfl.ch}}
\author{Paul Gafton, Erick Maraz}

\maketitle

\begin{abstract}
With the advent of Generative Adversarial Networks (GANs), a finer level of control in manipulating various features of an image has become possible. One example of such fine manipulation is changing the position of the light source in a scene. This is fundamentally an ill-posed problem, since it requires understanding the scene geometry to generate proper lighting effects. This problem is not a trivial one and can become even more complicated if we want to change the direction of the light source from any direction to a specific one.

Here we provide our attempt to solve this problem using GANs. Specifically, pix2pix\cite{pix2pix} trained with the dataset VIDIT\cite{vidit} which contains images of the same scene with different types of light temperature and 8 different light source positions (N, NE, E, SE, S, SW, W, NW).

The results are 8 neural networks trained to be able to change the direction of the light source from any direction to one of the 8 previously mentioned. Additionally, we provide, as a tool, a simple CNN trained to identify the direction of the light source in an image. \footnote{\url{https://github.com/Paul92/cp_2020}}

\end{abstract}

\section{Introduction}

In art, especially in photography, the light is a vital part of the artists' work. Digital tools have become an indispensable asset for a photographer, but their ability of manipulating light is limited to small variations in intensity or color. Artists still need to spend countless hours to find the appropriate light direction, with limited enhancements being possible in postprocessing. Our goal is to provide a tool that allows controlling the light direction in a scene.

To the best of our knowledge, there is no widely available tool for automatically changing the lighting in an image. There has been some academic research in developing such a system \cite{3dScene}, \cite{deepface}, but its applications are somewhat limited. Ideally, we would be interested in a method that allows the change of light source position in a 2D image, without any additional input, such as geometric priors, depth information or the current position of the light source.

One of the most important cues of the light source position are the shadows. \cite{3dScene} produces one of the most visually appealing results for large scale scenes, but requires a 3D model of the scene for determining the current shadows and casting shadows from the new light direction. These shadow maps are then processed along with the image by a relighting network that produces the final result. \cite{deepface} has recently tackled image relighting without an explicit geometric prior, requiring only the image itself as input. Instead, they relied on the relatively regular geometry of the human face.

In more general terms, consistent image manipulations, such as style transfer, have recently been made available to the public, both for artistic purposes and for entertainment. However, the goal of these techniques is only to replicate the overall style of an image, with no control over individual details. With the evolution of neural networks, especially GANs, it has become possible to control individual aspects of an image, while preserving global consistency. For instance, CycleGAN \cite{cycleGAN} demonstrated a horse to zebra transformation using a cycle consistency loss and StyleGAN \cite{styleGAN}, \cite{styleGAN2} portrayed the gradual transfer of features from one face to another. Other publications, such as \cite{semanticHierarchy}, attempts changing different aspects of a room by manipulating the latent space of a GAN. Considering the performance of the GAN-based applications for global image transformations, we propose discarding the geometric prior and treating the relighting problem as an image to image translation problem.

The following section presents a literature review, starting from the state-of-the-art approaches to illumination changes and continuing with the style transfer and the proposed image-to-image translation techniques.

\section{Related Work}

The shadows are the result of the interaction between the light and the scene geometry. Therefore, a common approach to the image relighting problem is to explicitly use the scene geometry as a prior of the algorithm.

\cite{duchene} demonstrates outdoor scene relighting by removing the present shadows and recasting them using a 3D geometric prior. Based on this, \cite{3dScene} achieves state-of-the-art results in image relighting. To this end, their proposed pipeline starts from a 3D geometric prior, that can be generated using publicly available tools for multiview geometry, such as \cite{realityCapture} and \cite{meshRoom}. A good part of the work in \cite{3dScene} is generating accurate shadow maps, whose importance has been shown in previous works as well \cite{duchene}. While \cite{duchene} formulates the shadow segmentation problem using a Markov Random Field over a graph of points, \cite{3dScene} directly casts the shadows from the 3D prior, leveraging the camera calibration from the multiview geometry reconstruction. The possible errors in the 3D prior affect the quality of the shadow maps, which are corrected by shadow refinement networks, one for the original shadows and one for the target shadows. Besides the shadow maps, the relighting network leverages the geometrical information of the scene through normal maps and reflection buffers. Both, the relighting network and the shadow refinement networks, use ResNet architectures \cite{resnet}, the authors reporting only marginally lower results with U-Net architectures \cite{unet}.

Another approach to relighting is taken by \cite{deepface}, which is the-state-of-the-art in portrait relighting. Leveraging the relative uniform geometry of the human faces and the extensive research in detecting faces and face landmarks, \cite{deepface} accurately retrieves the surface normals required for relighting from a single image. More specifically, they use SfSNet \cite{sfsnet} for estimating the Spherical Harmonics lighting and a U-Net architecture where the SH lighting parameters are manipulated in the bottleneck layer.

Since the change in illumination can produce significant changes in the appearance of a scene, this can limit the performance of deep learning approaches. \cite{roadRelight} demonstrates an application of image relighting for data augmentation in autonomous driving problems, such as segmentation or object detection. In this case, the geometric information is extracted from the depth map and combined with the information from road segmentation. The luminance and the color components of the output are computed separately, by two hourglass networks, where the light direction is added in bottleneck layer, similar to \cite{deepface}.

The image-to-image translation is a class of vision and graphics problems where the goal is to learn a mapping between the input image and the output image, usually for applying a non-trivial change in a globally consistent manner. Many vision and image processing problems can be formulated as image to image translation problems, starting from the basic edge detection, up to more complex applications such as style transfer or image segmentation. So far, most of these challenges have been solved with hand crafted solutions, either using traditional algorithms or through specially designed machine learning models. The problem with most these of the machine learning models is the difficulty in designing a good loss function - usually the desired outcome is not to exactly replicate the training data, but to produce a globally consistent and visually appealing result. This requirement makes GANs \cite{theGAN} an ideal candidate for this type of problems, since they can infer the loss from the training data, with works such as pix2pix \cite{pix2pix}, CycleGan \cite{cycleGAN} successfully employing GANs in image-to-image translation problems.

Instead of training the generator model using an error function, the fundamental concept behind GANs is to use another model during training, called discriminator. The generator and the discriminator are trained alternatively, competing in a zero sum game: the generator learns to produce data more and more alike the training data and the discriminator learns to find the mistakes of the generator. Traditionally, the input of the generator is a vector, but the exact way this is mapped to the output is far from trivial and is under active research \cite{outputMap1}, \cite{outputMap2}. One way of having better control over the output of the generator is to condition its output, with a type of models known as Conditional GANs \cite{conditionalGAN}. In Conditional GANs, instead of learning a mapping from the latent space to the output, the generator learns a mapping conditioned by some additional input. One type of such conditioning can be an input image. This idea has been successfully used in many applications \cite{conditionalGANapplication}, but, to the best of our knowledge, pix2pix \cite{pix2pix} is the first framework that introduced a general solution to the image-to-image translation problem, and at the same time providing enough flexibility to handle a large palette of cases. One limitation of pix2pix is the requirement of paired data. To address this constraint, CycleGAN \cite{cycleGAN} introduces the concept of cycle consistency in GANs by training not only the forward mapping, but also an inverse mapping and introducing a loss based on their composition.

\section{Proposed Method}

Using hourglass-shaped networks for relighting tasks is common practice across literature \cite{deepface}, \cite{3dScene}, \cite{roadRelight}. We demonstrate the capacity of such networks to produce a full image relighting, without any additional help by introducing an image-to-image translation solution. Since one network can provide only a single target direction, we use an ensemble of eight identical models for changing the light source position in one of the eight available target positions, regardless of the initial light direction. Thus, our approach results in a forty-five-degree accuracy of the light source direction, at a constant light elevation.

Based on pix2pix \cite{pix2pix}, all the models follow a U-Net-256 architecture. 

It has been observed that for image generation tasks, the L1 and L2 losses model very well the low frequencies, but fail at capturing the high frequencies, thus leading to blurry images. In order to model the high frequencies, we used a patch-based discriminator, as suggested by pix2pix \cite{pix2pix}. This discriminator is ran convolutionally across the image and the responses are averaged to produce the final discriminator output. The discriminator used has an input size of 70x70.

\autoref{eq:loss_gan}, \autoref{eq:l1_reg} and \autoref{eq:objective} show the loss and objective functions used in the model where the conditional GAN learns a mapping from, an observed image $x$ with random noise $z$, to $y$, $G:\{x,z\} \rightarrow y$. See \cite{pix2pix} for the complete explanation and derivation of the model. 
\begin{equation}
    \mathcal{L}_{cGAN}\left(G, D\right) = \mathbb{E}_{x,y} \left[\log \left( D\left(x,y\right)\right)\right]+ \mathbb{E}_{x,z} \left[\log\left(1-D\left(x, G\left(x, z\right)\right)\right)\right]
    \label{eq:loss_gan}
\end{equation}
\begin{equation}
\mathcal{L}_{L1}(G)=\mathbb{E}_{x,y,z}\left[\left\lVert y - G\right(x,z\left) \right\rVert_1\right]
    \label{eq:l1_reg}
\end{equation}
\begin{equation}
    G^{*} = \arg \min_G \max_D \mathcal{L}_{cGAN}\left(G,D\right)+\lambda \mathcal{L}_1\left(G\right)
    \label{eq:objective}
\end{equation}

In order to generalize to an illumination transfer method, we trained the models paring them in the following way: each different direction to only one direction. This, with the goal of somehow tell the GAN what are we trying to do.

\begin{figure}[H]
\centering
    \includegraphics[scale=0.55]{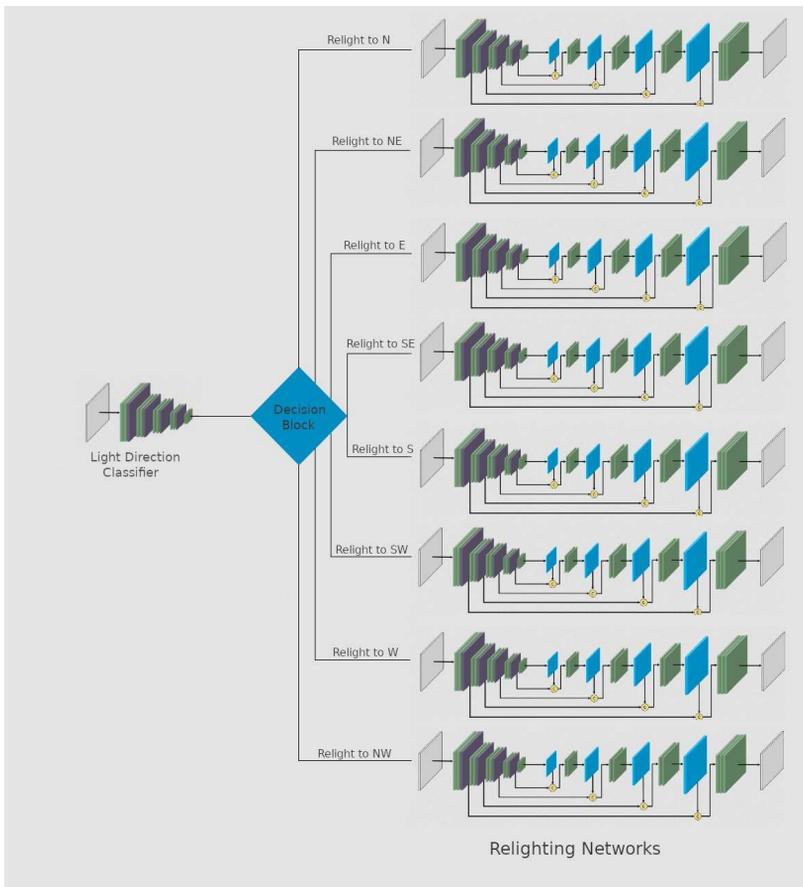}
    \caption{Full architecture of the scene relighting network}
    \label{fig:architecture}
\end{figure}

Besides that, we use a classifier for determining the source-light position and choose the appropriate model to apply, as shown in \autoref{fig:architecture}. Determining the direction of the light is a widely studied problem in Augmented Reality \cite{AR1}, \cite{AR2}, where it is important to shade correctly the introduced objects.

\begin{figure}[h!]
\centering
    \includegraphics[scale=0.6]{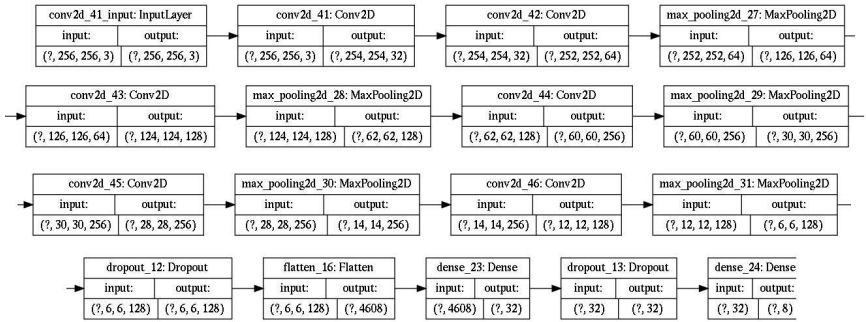}
    \caption{Light direction classifier architecture: a CNN with convolutional and pooling layers}
    \label{fig:classifier}
\end{figure}

We have experimented with a simple approach based on a CNN classifier for determining the light direction out of the eight possible. The architecture of this model is described in \autoref{fig:classifier}.

%\vspace{1em}

The full architecture of our approach is therefore shown in \autoref{fig:architecture}.

%\vspace{1em}

\section{Experimental Results}

Our method is composed of two main parts: the relighting networks and the light direction classifier. We start by discussing the dataset, after which we will provide an in-depth analysis of the most important component, the relighting networks, looking specifically at their performance and evolution during training. Following this discussion, we will analyse the light direction classifier.

\subsection{Data}

The VIDIT \cite{vidit} dataset is a synthetically generated dataset of images from multiple scenes, where the light source is varied in 8 different positions with respect to the camera, with 5 different temperatures. The main advantage of using a synthetic dataset is the accuracy of the ground truth, as well as the ability to address the simplified problem of image relighting, using scenes with only one light source.

To test the relighting capabilities, we picked images at a constant temperature of 4500K, and preprocessed them in the format required by the pix2pix \cite{pix2pix} framework. The train-test split was about 90\%-10\%, with randomly chosen images.

\subsection{Relighting Networks}

\begin{figure}[H]
\centering
    \includegraphics[scale=0.5]{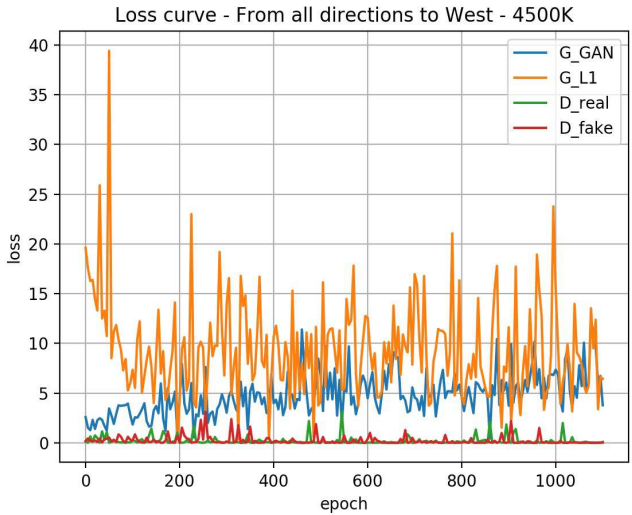}
    \caption{Loss curve - \textbf{4500K-W} }
    \label{fig:loss_epoch}
\end{figure}

By comparing visually the outputs of the relighting networks, we can make some observations regarding their performance. In most cases, the general light direction and shadows are reasonably well determined. There are, however, great limitations for recovering dark shadows - we assume this is due to the very little amount of detail, most of the scenes lacking indirect or secondary lighting. There are also some issues with preserving complex, artificial scenes, which contain a great number of straight lines. Even tough, we tried different batch sizes and various hyper parameters, the best results were obtained with the default values of the pix2pix framework. We generated periodically samples and looked at them to find the best hyper parameters. One example of the loss curves we obtained is shown in \autoref{fig:loss_epoch} where $G$ stands for generator and $D$ for discriminator. The loss curves for the other cases show a similar behaviour. 

\begin{figure}[H]
\centering

\begin{subfigure}{0.32\textwidth}
\centering
\includegraphics[width=0.8\linewidth]{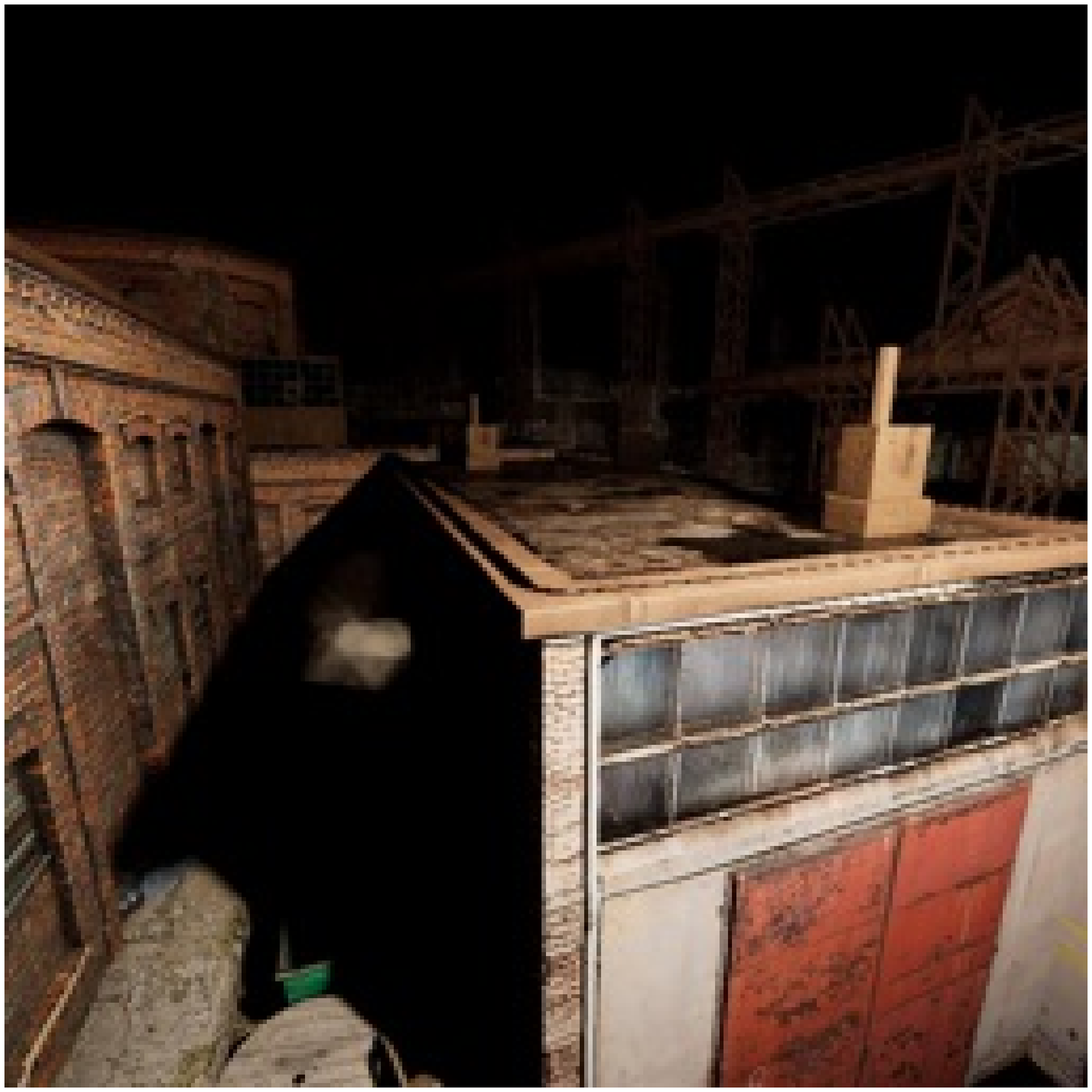} 
\label{fig:subim1}
\end{subfigure}
\begin{subfigure}{0.32\textwidth}
\centering
\includegraphics[width=0.8\linewidth]{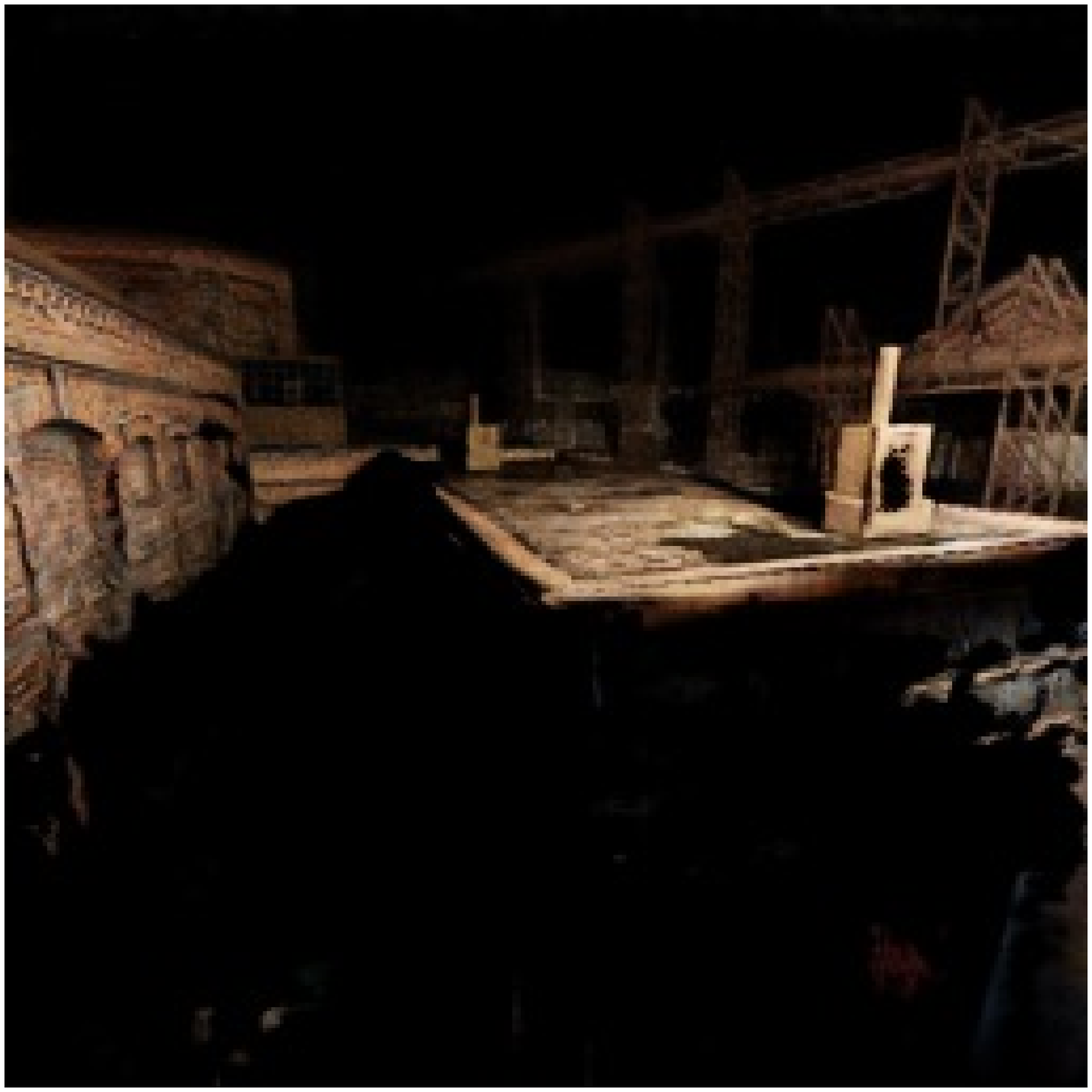} 
\label{fig:subim2}
\end{subfigure}
\begin{subfigure}{0.32\textwidth}
\centering
\includegraphics[width=0.8\linewidth]{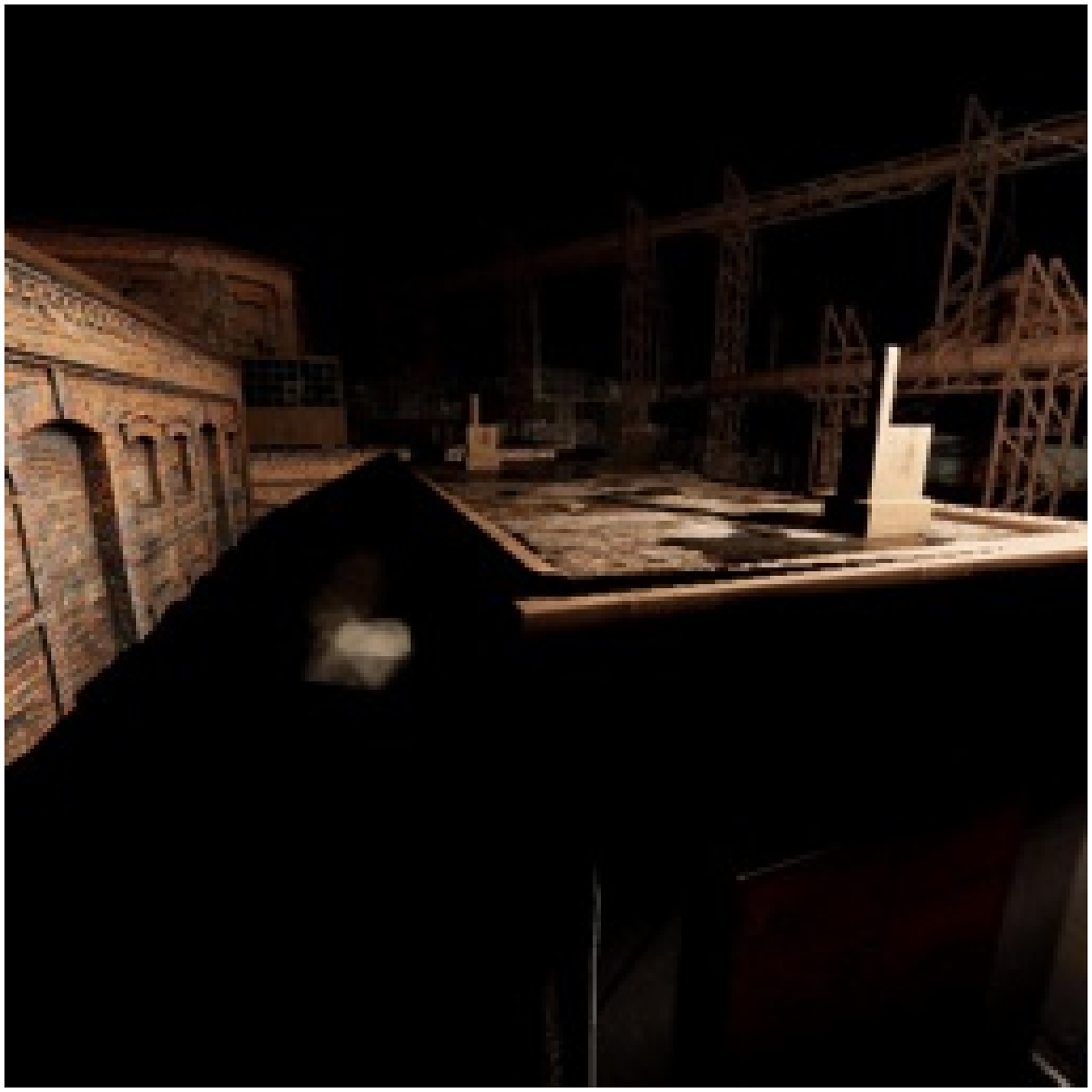} 
\label{fig:subim2}
\end{subfigure}

\begin{subfigure}{0.32\textwidth}
\centering
\includegraphics[width=0.8\linewidth]{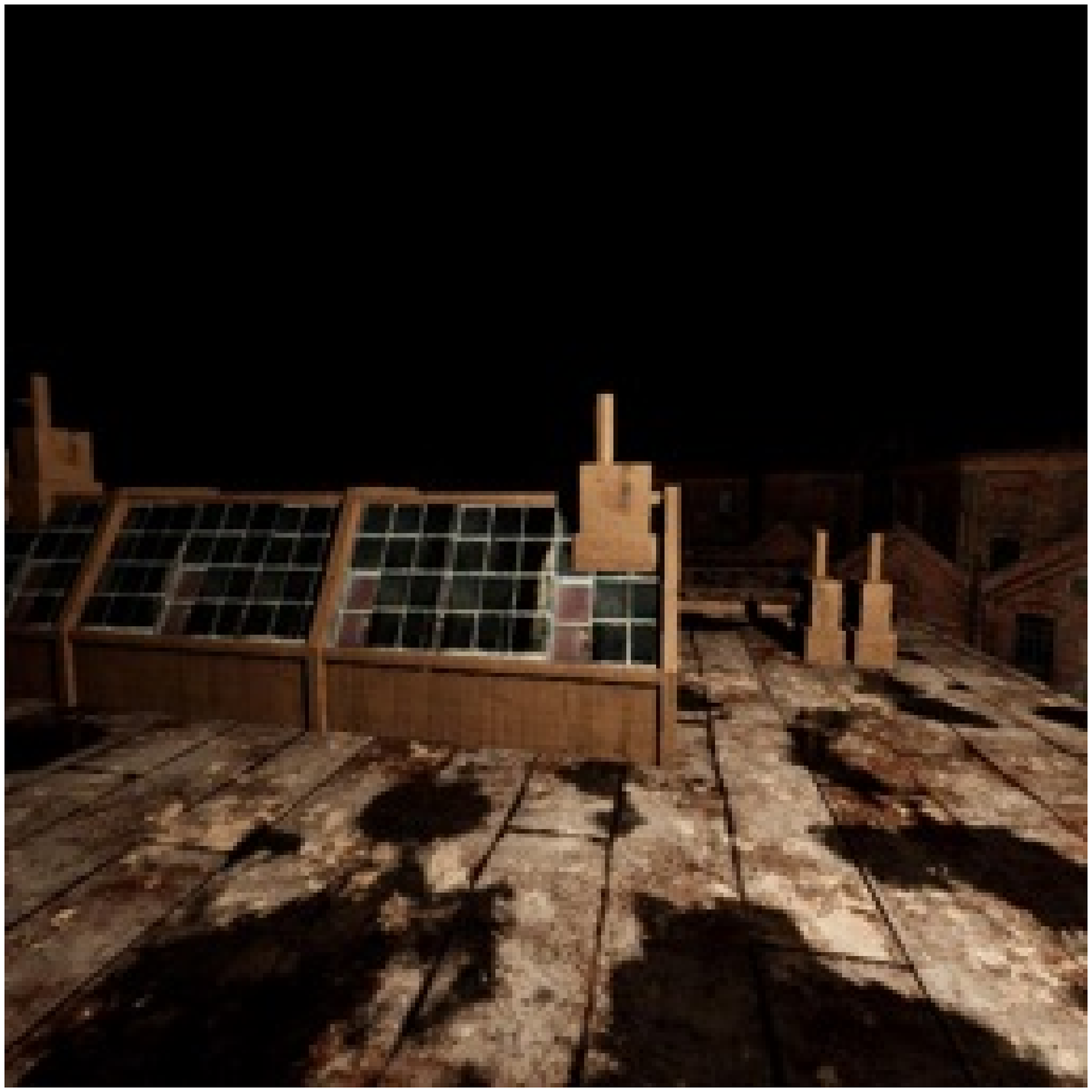} 
\label{fig:subim1}
\end{subfigure}
\begin{subfigure}{0.32\textwidth}
\centering
\includegraphics[width=0.8\linewidth]{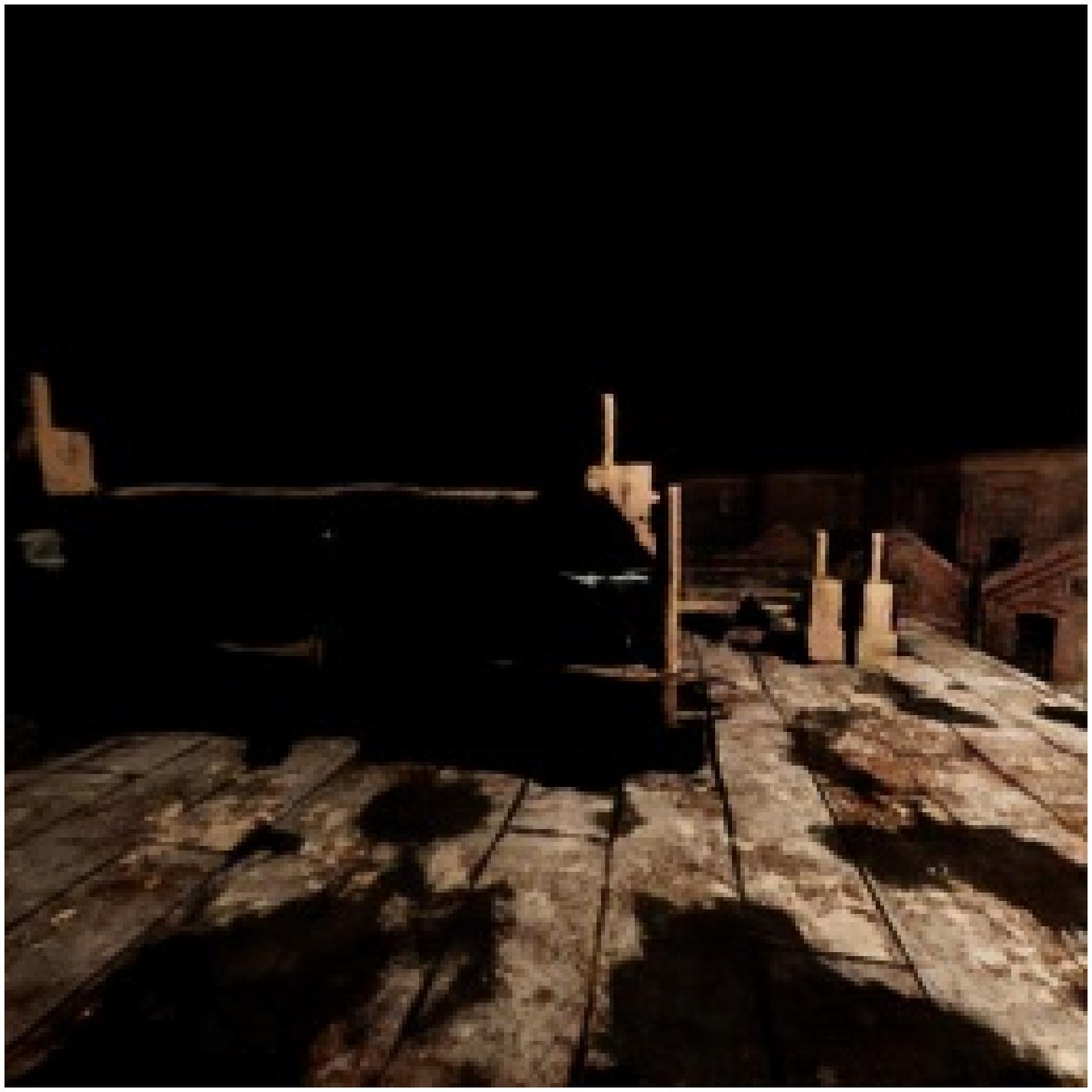} 
\label{fig:subim2}
\end{subfigure}
\begin{subfigure}{0.32\textwidth}
\centering
\includegraphics[width=0.8\linewidth]{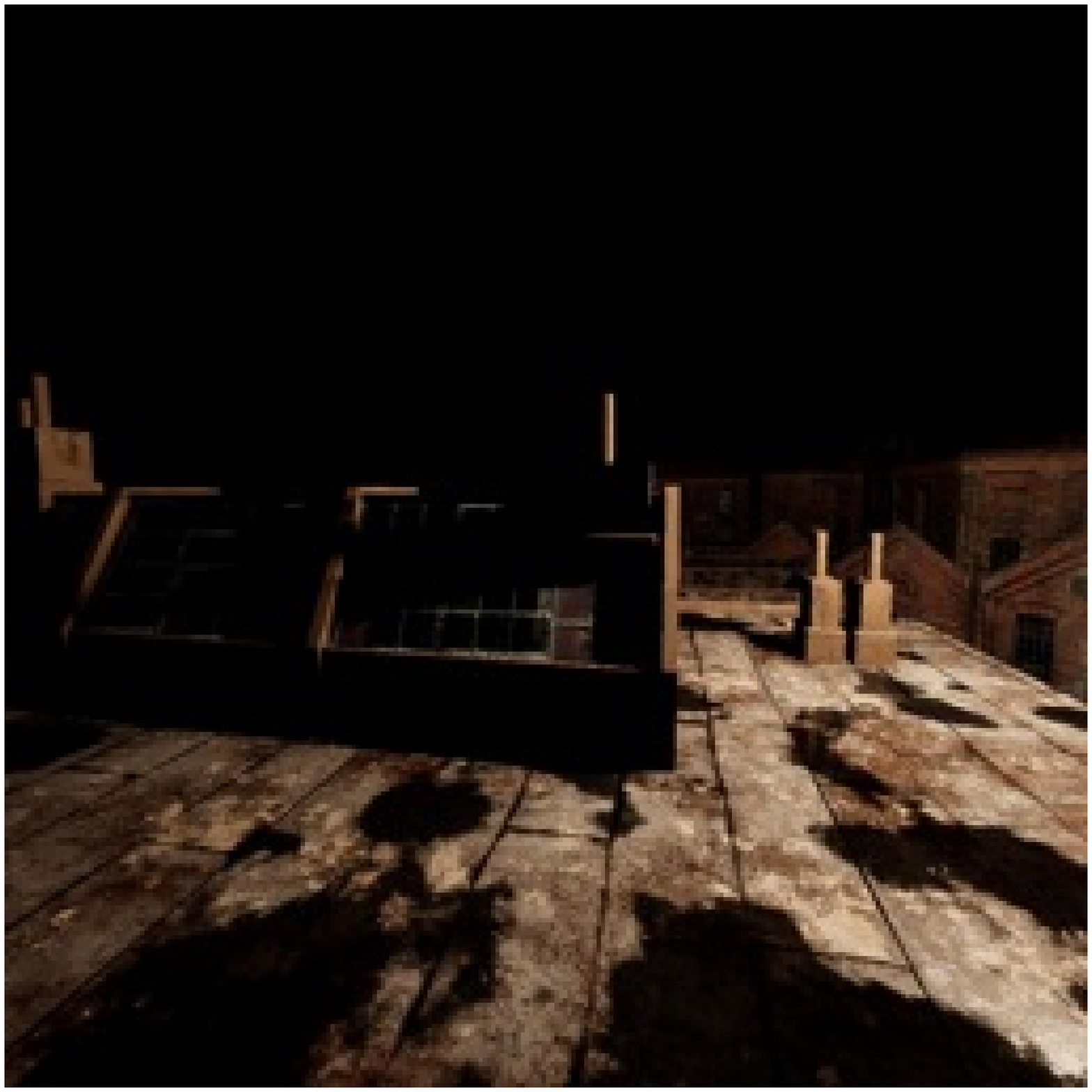} 
\label{fig:subim2}
\end{subfigure}

\begin{subfigure}{0.32\textwidth}
\centering
\includegraphics[width=0.8\linewidth]{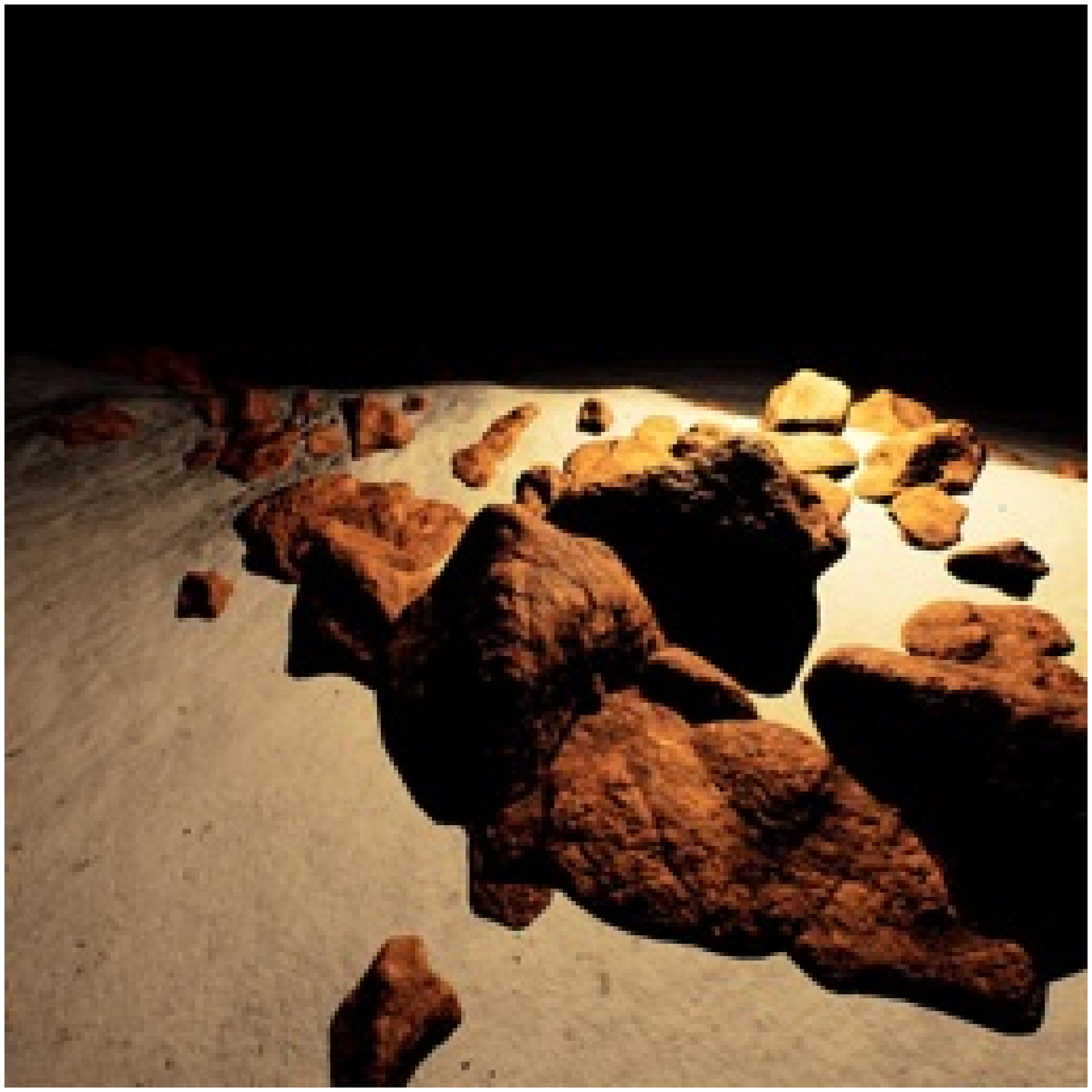} 
\label{fig:subim1}
\end{subfigure}
\begin{subfigure}{0.32\textwidth}
\centering
\includegraphics[width=0.8\linewidth]{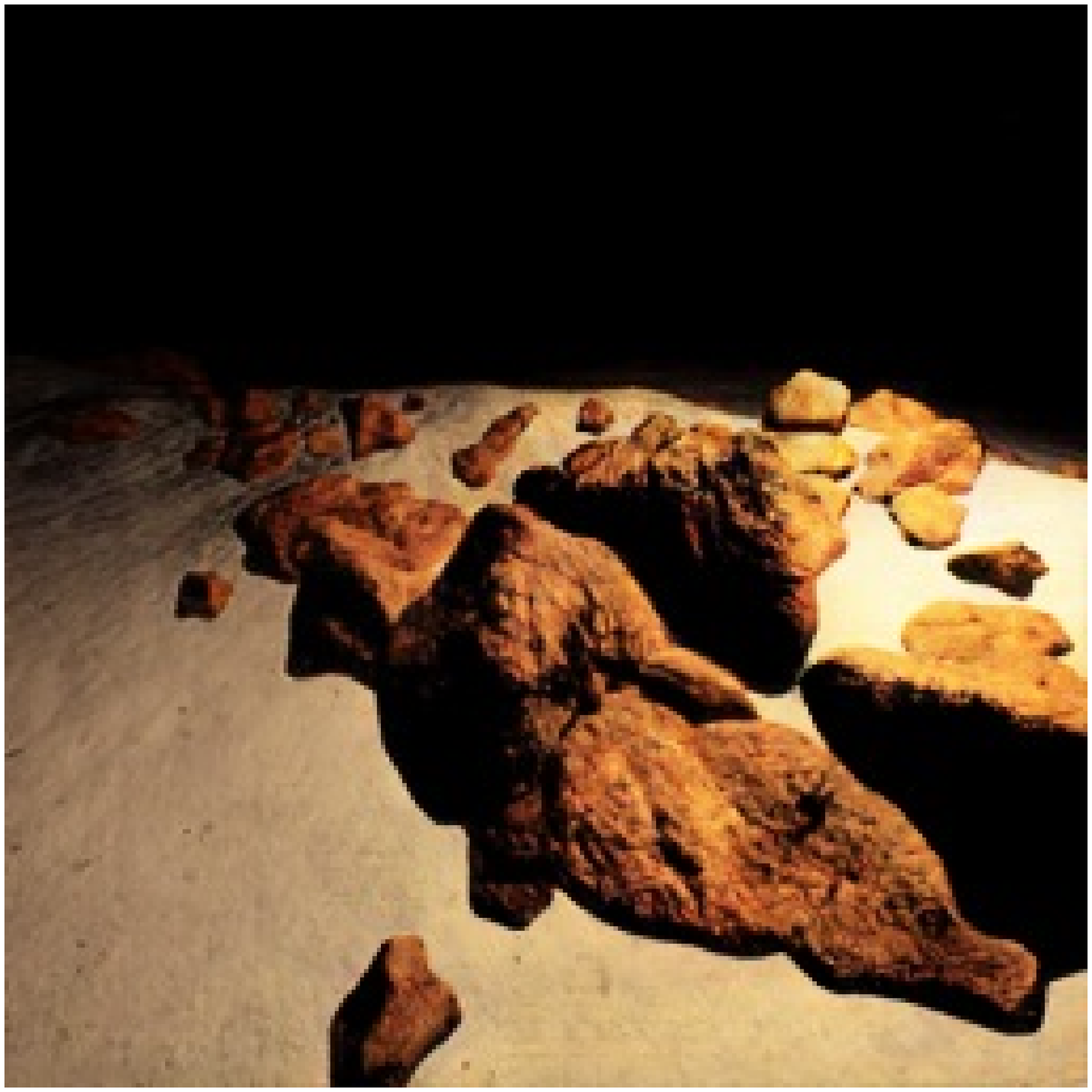} 
\label{fig:subim2}
\end{subfigure}
\begin{subfigure}{0.32\textwidth}
\centering
\includegraphics[width=0.8\linewidth]{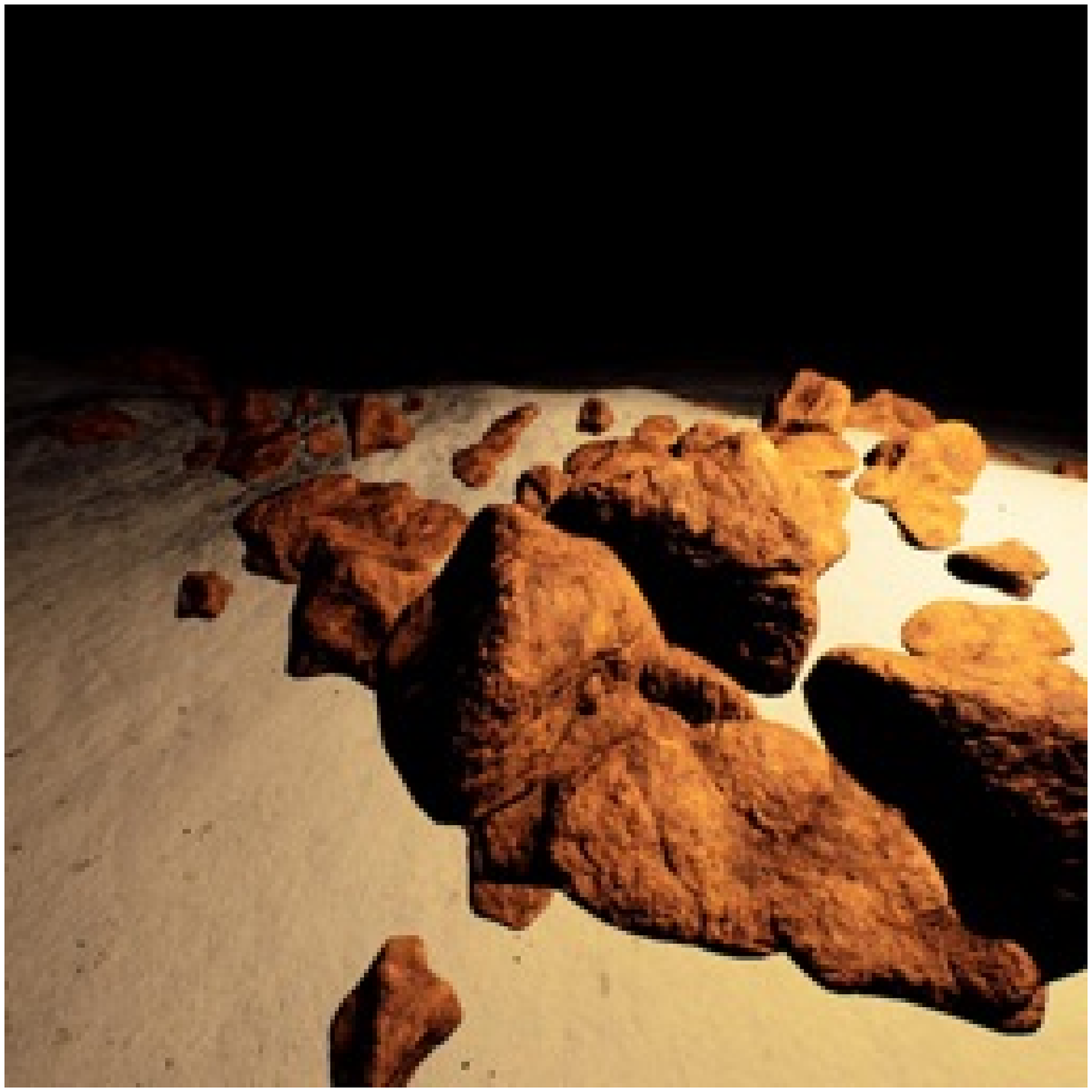} 
\label{fig:subim2}
\end{subfigure}

\begin{subfigure}{0.32\textwidth}
\centering
\includegraphics[width=0.8\linewidth]{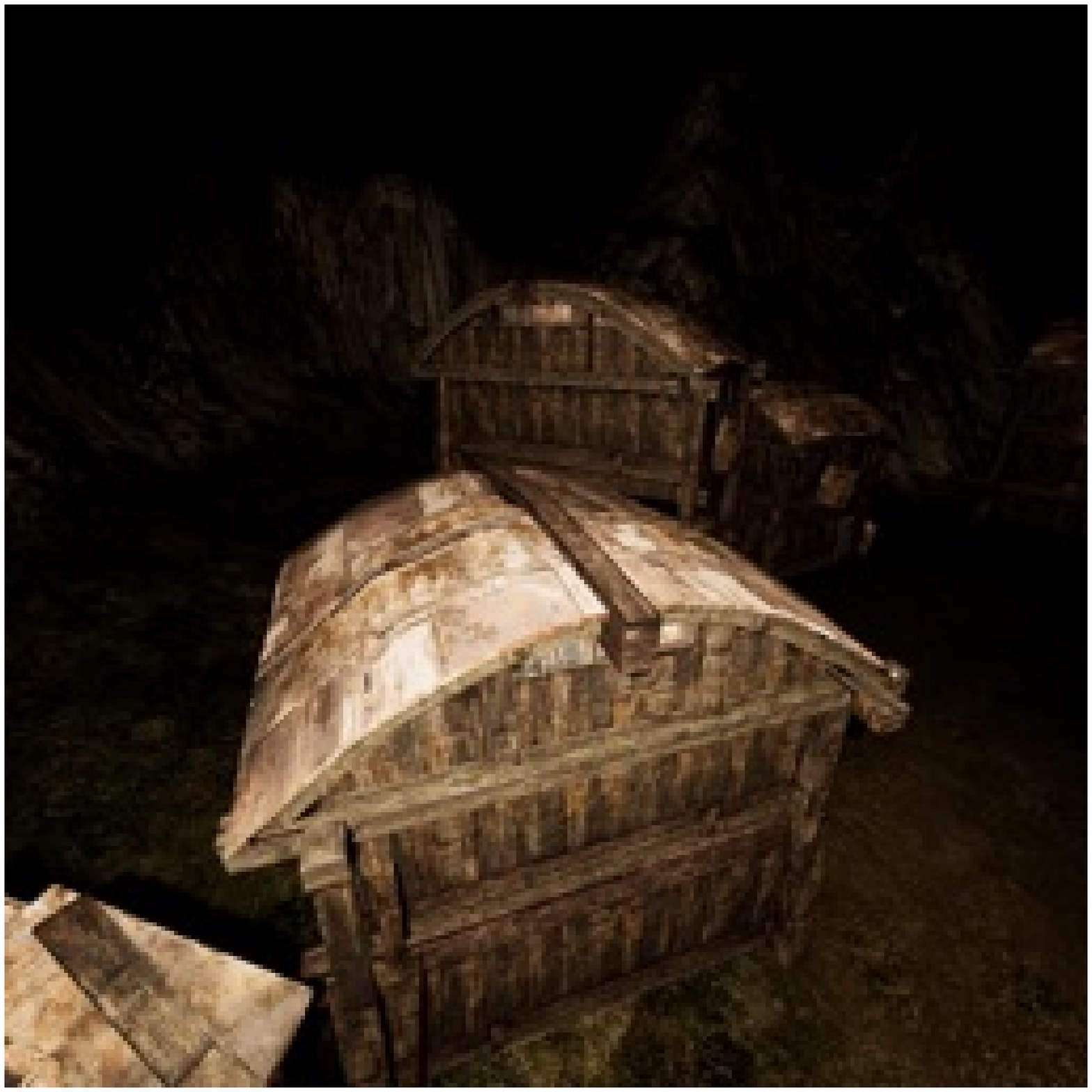} 
\label{fig:subim1}
\end{subfigure}
\begin{subfigure}{0.32\textwidth}
\centering
\includegraphics[width=0.8\linewidth]{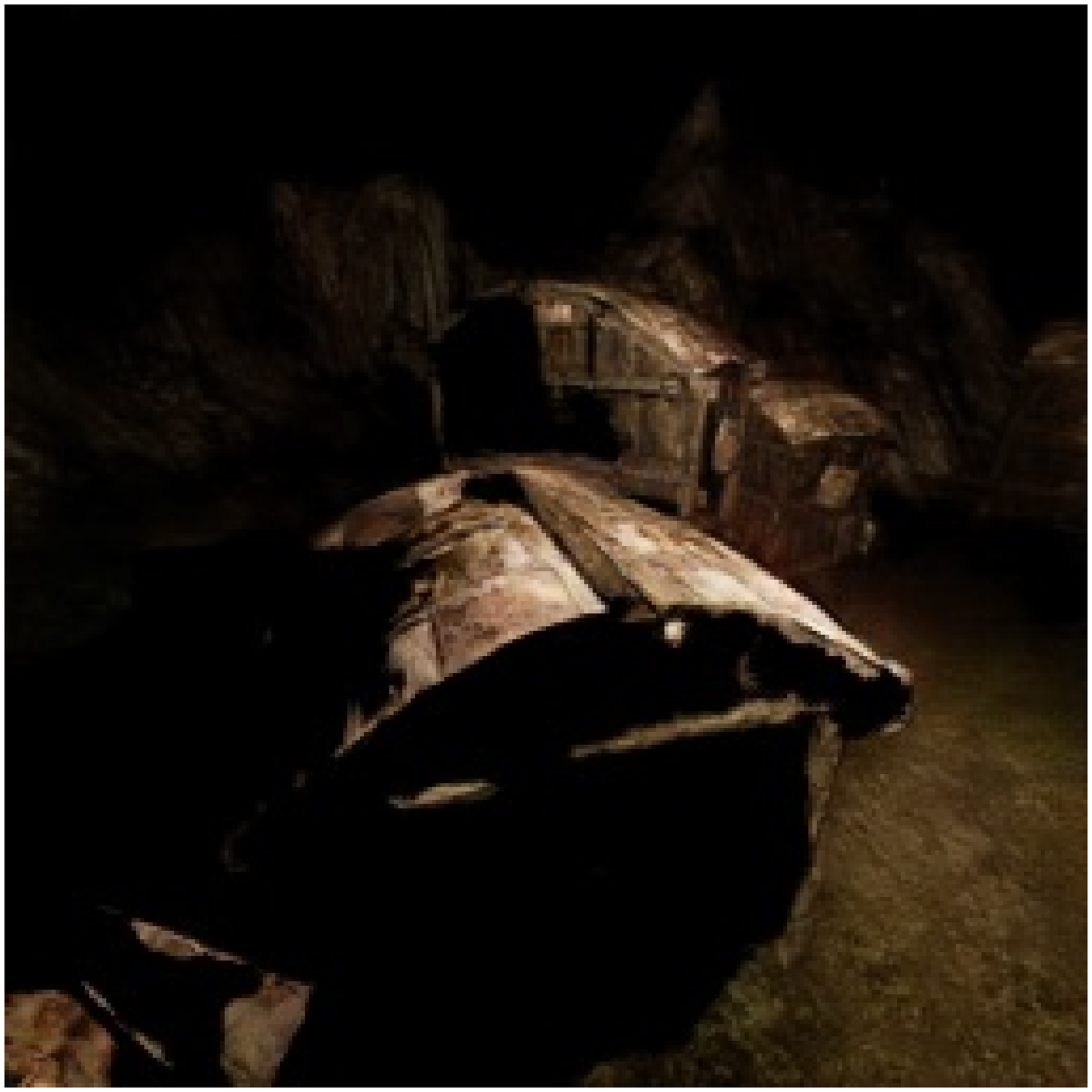} 
\label{fig:subim2}
\end{subfigure}
\begin{subfigure}{0.32\textwidth}
\centering
\includegraphics[width=0.8\linewidth]{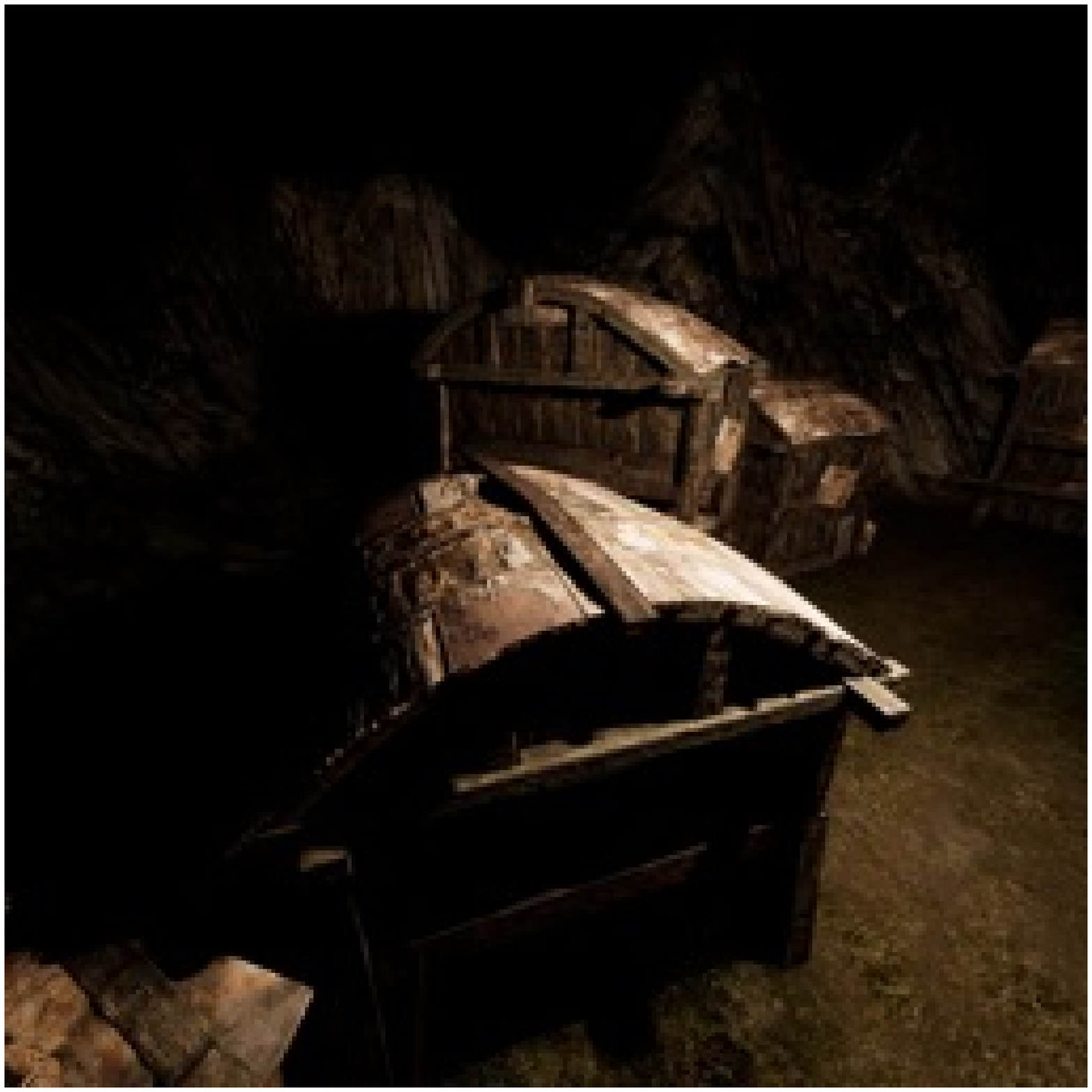} 
\label{fig:subim2}
\end{subfigure}

\begin{subfigure}{0.32\textwidth}
\centering
\includegraphics[width=0.8\linewidth]{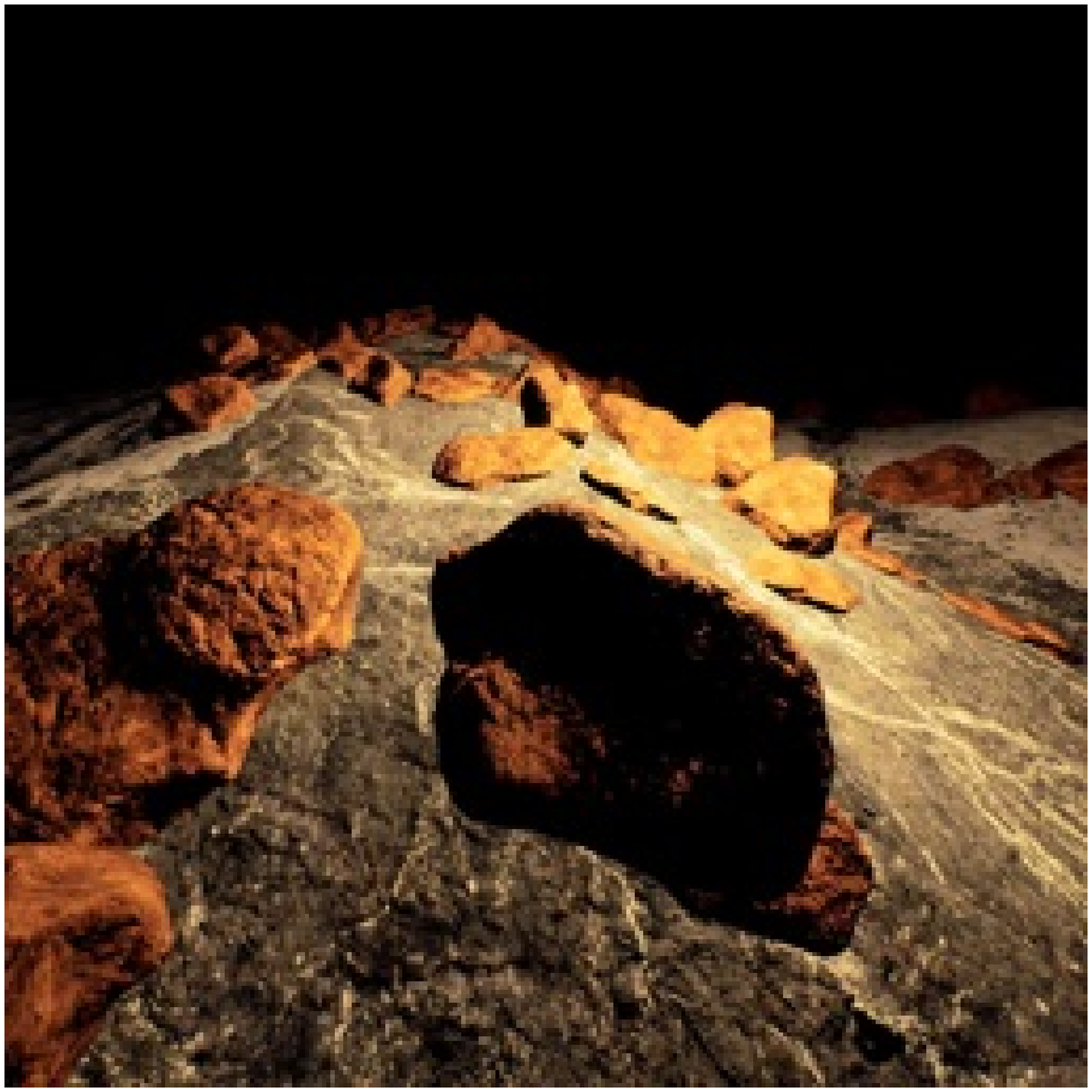} 
\caption{Input}
\label{fig:subim1}
\end{subfigure}
\begin{subfigure}{0.32\textwidth}
\centering
\includegraphics[width=0.8\linewidth]{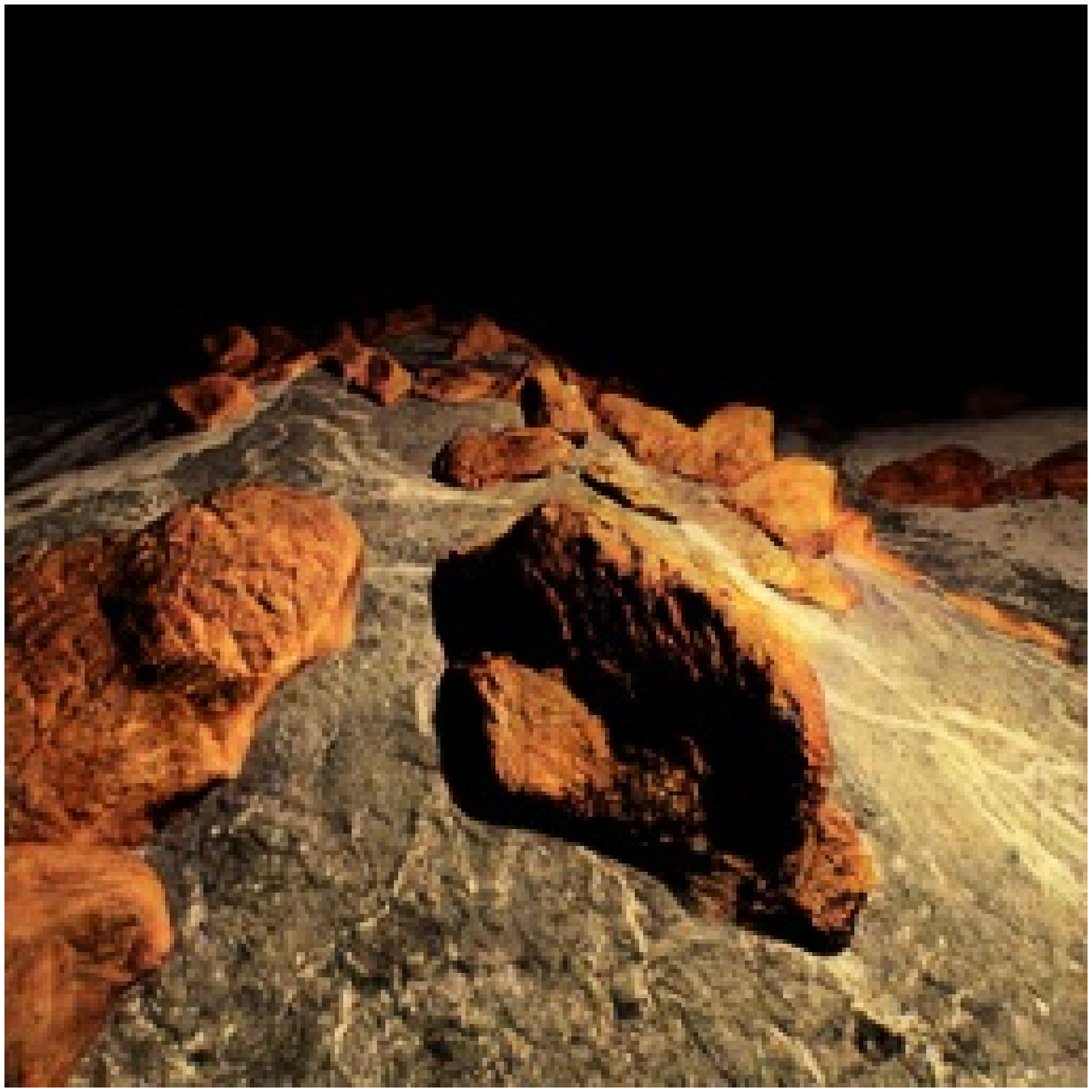} 
\label{fig:subim2}
\caption{Output}
\end{subfigure}
\begin{subfigure}{0.32\textwidth}
\centering
\includegraphics[width=0.8\linewidth]{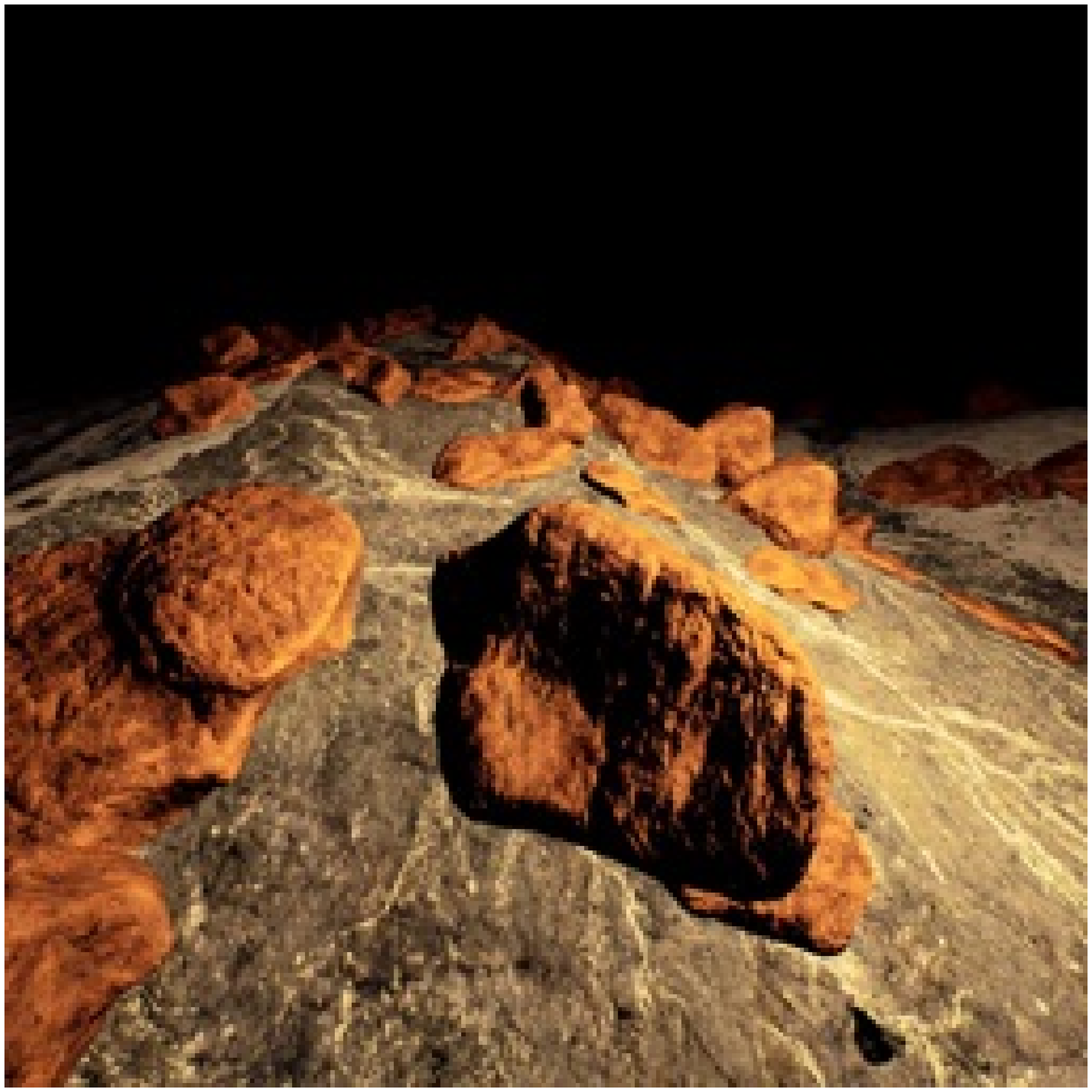} 
\label{fig:subim2}
\caption{Ground Truth}
\end{subfigure}

\caption{Successfully relit scenes, with very little distortion and accurate shadow casting}
\label{fig:relight_net}
\end{figure}

\begin{figure}[H]
\centering

\begin{subfigure}{0.32\textwidth}
\centering
\includegraphics[width=0.8\linewidth]{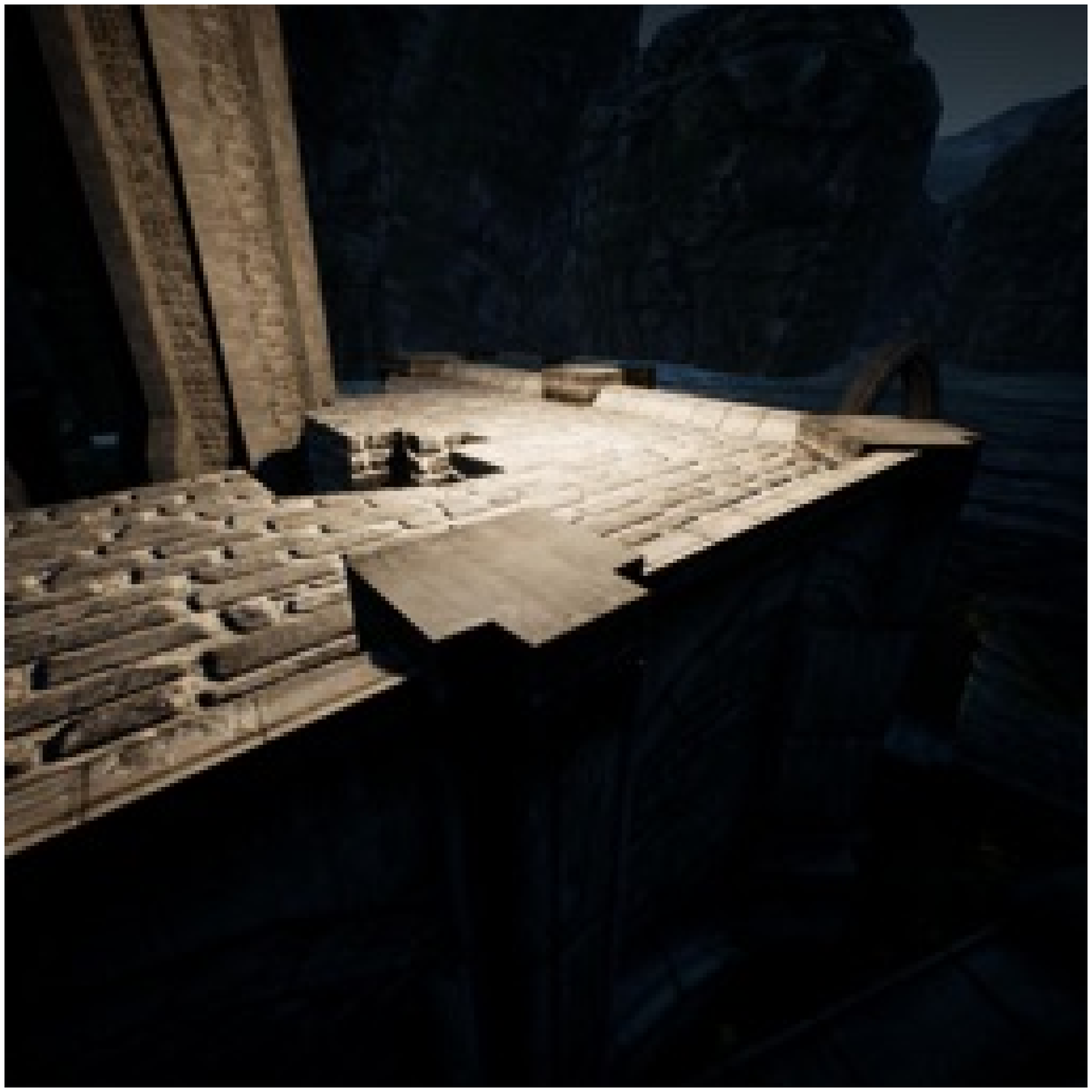} 
\label{fig:subim1}
\end{subfigure}
\begin{subfigure}{0.32\textwidth}
\centering
\includegraphics[width=0.8\linewidth]{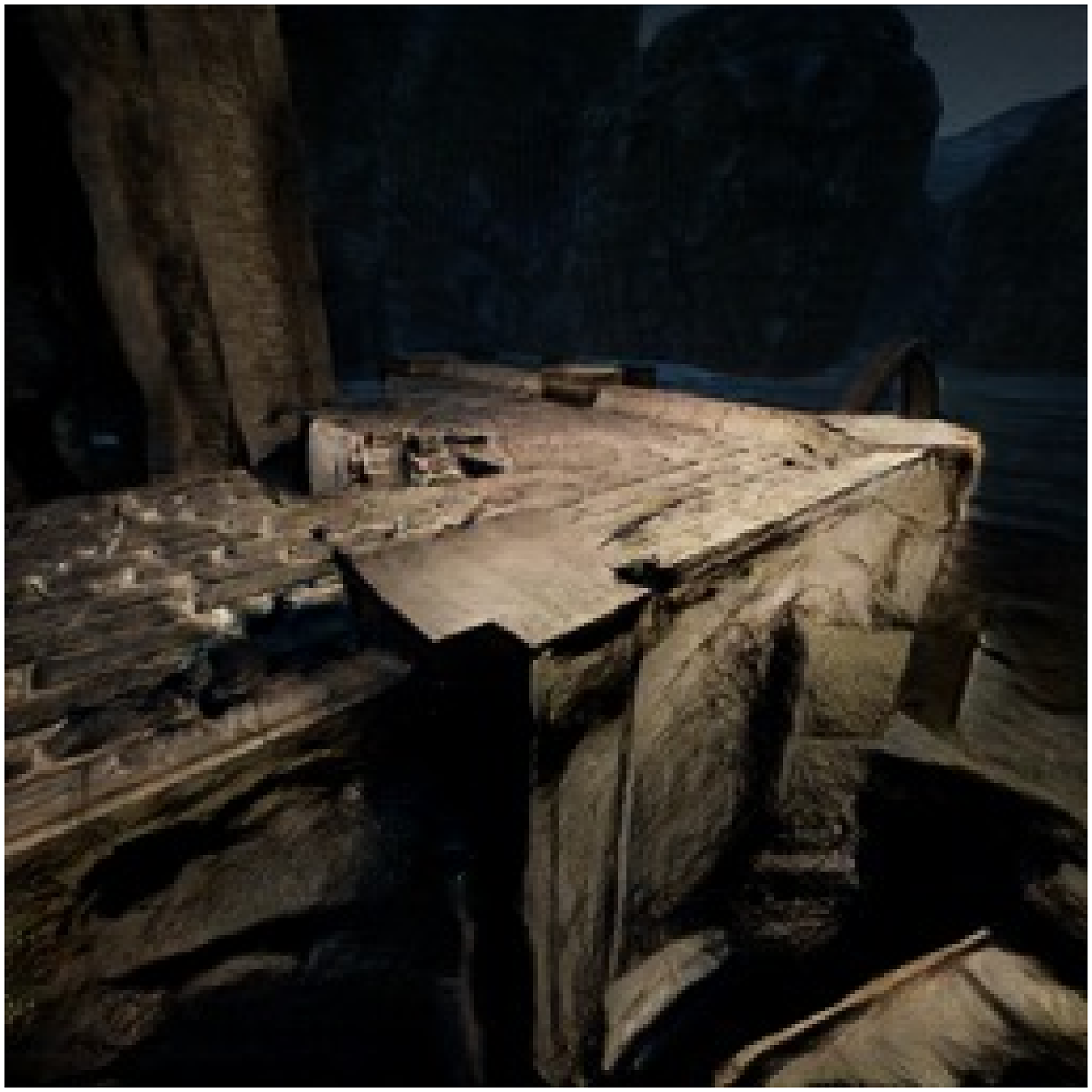} 
\label{fig:subim2}
\end{subfigure}
\begin{subfigure}{0.32\textwidth}
\centering
\includegraphics[width=0.8\linewidth]{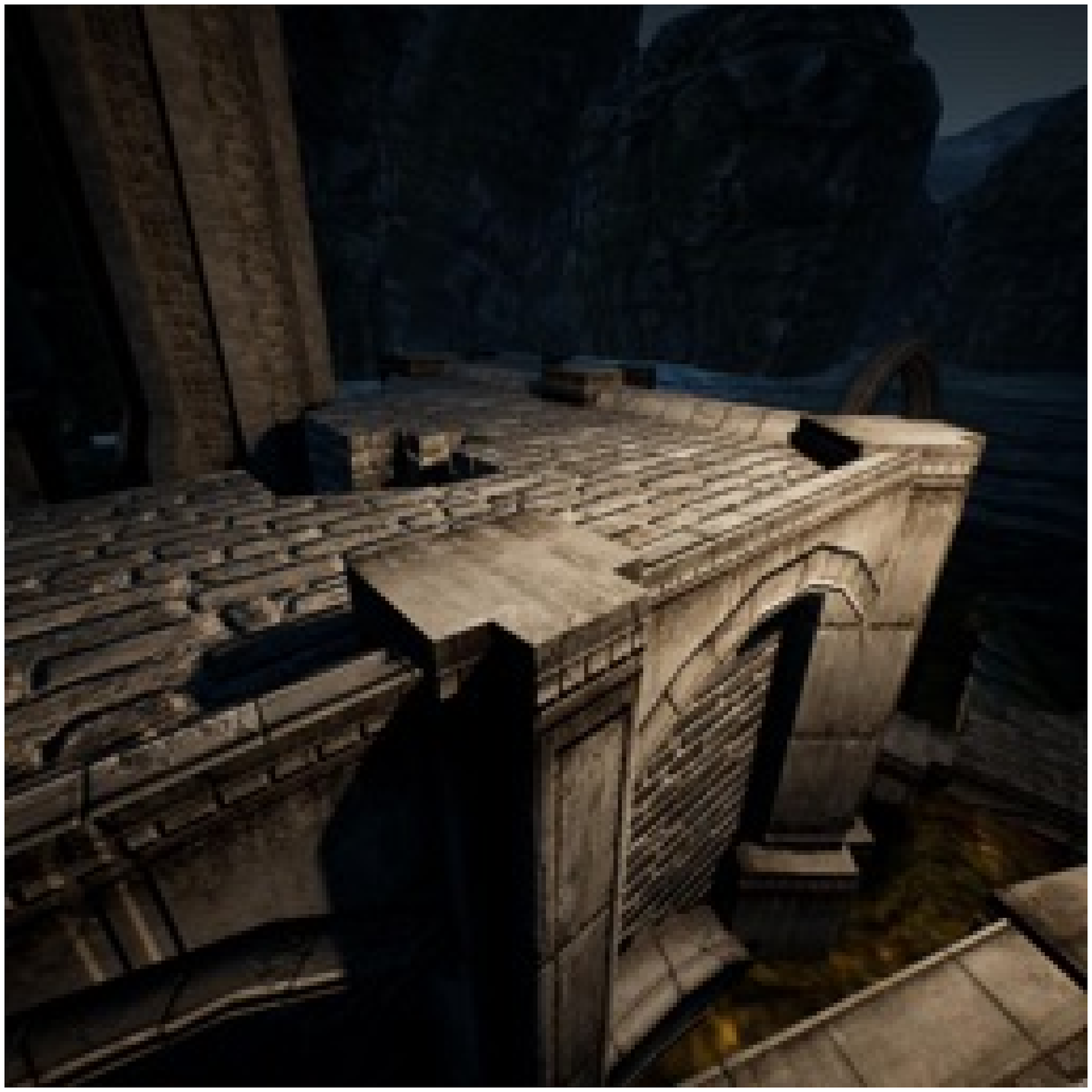} 
\label{fig:subim2}
\end{subfigure}

\begin{subfigure}{0.32\textwidth}
\centering
\includegraphics[width=0.8\linewidth]{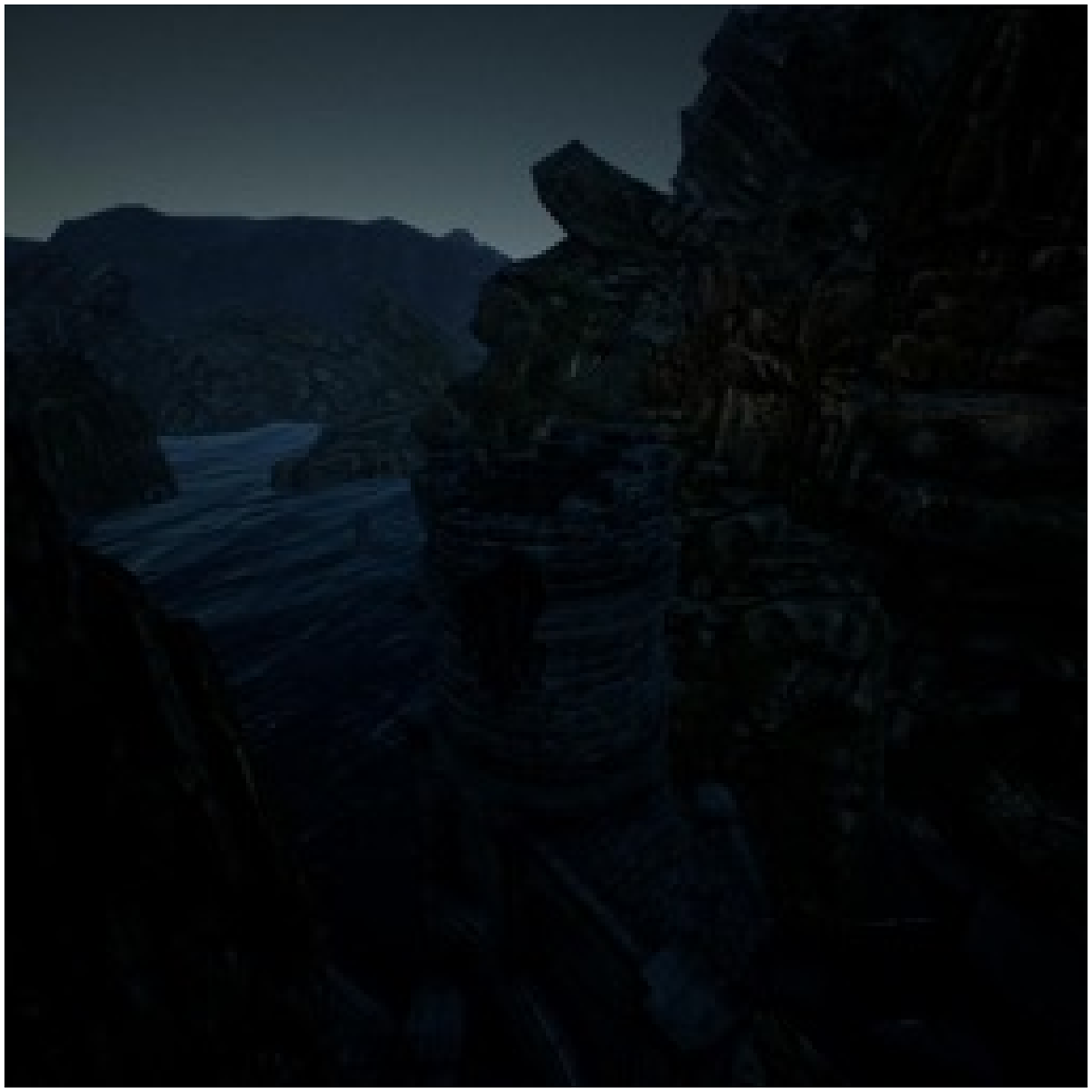} 
\label{fig:subim1}
\end{subfigure}
\begin{subfigure}{0.32\textwidth}
\centering
\includegraphics[width=0.8\linewidth]{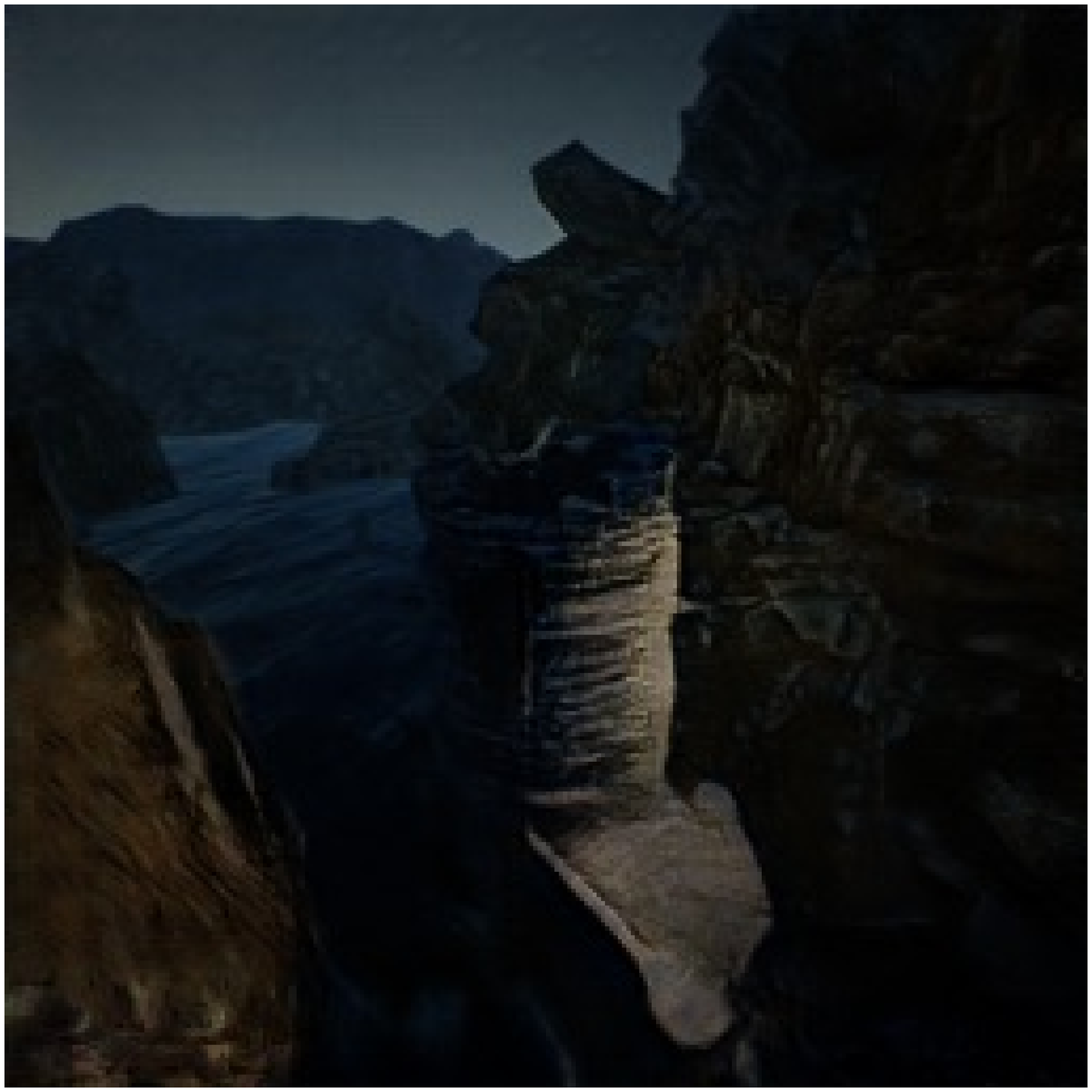} 
\label{fig:subim2}
\end{subfigure}
\begin{subfigure}{0.32\textwidth}
\centering
\includegraphics[width=0.8\linewidth]{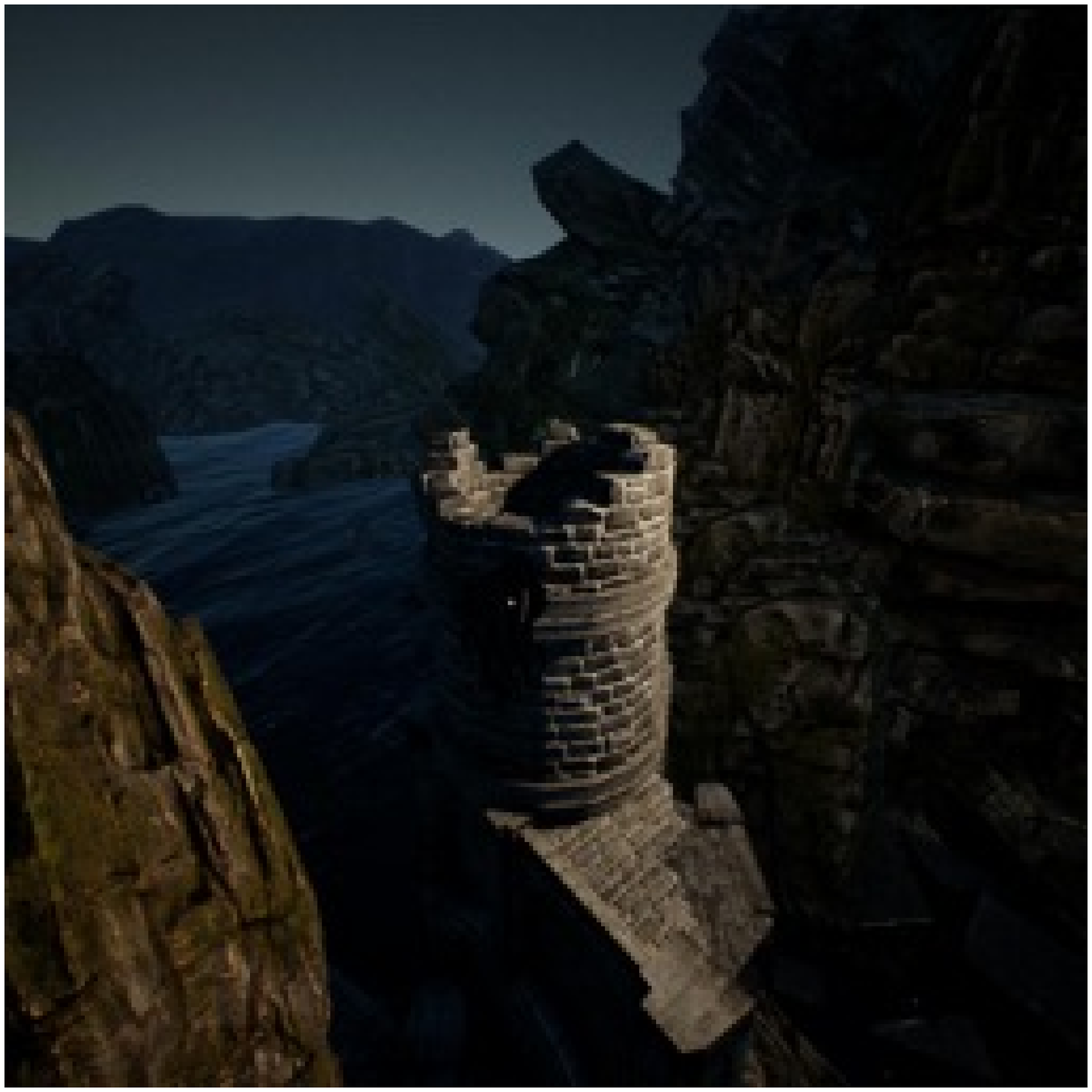} 
\label{fig:subim2}
\end{subfigure}

\begin{subfigure}{0.32\textwidth}
\centering
\includegraphics[width=0.8\linewidth]{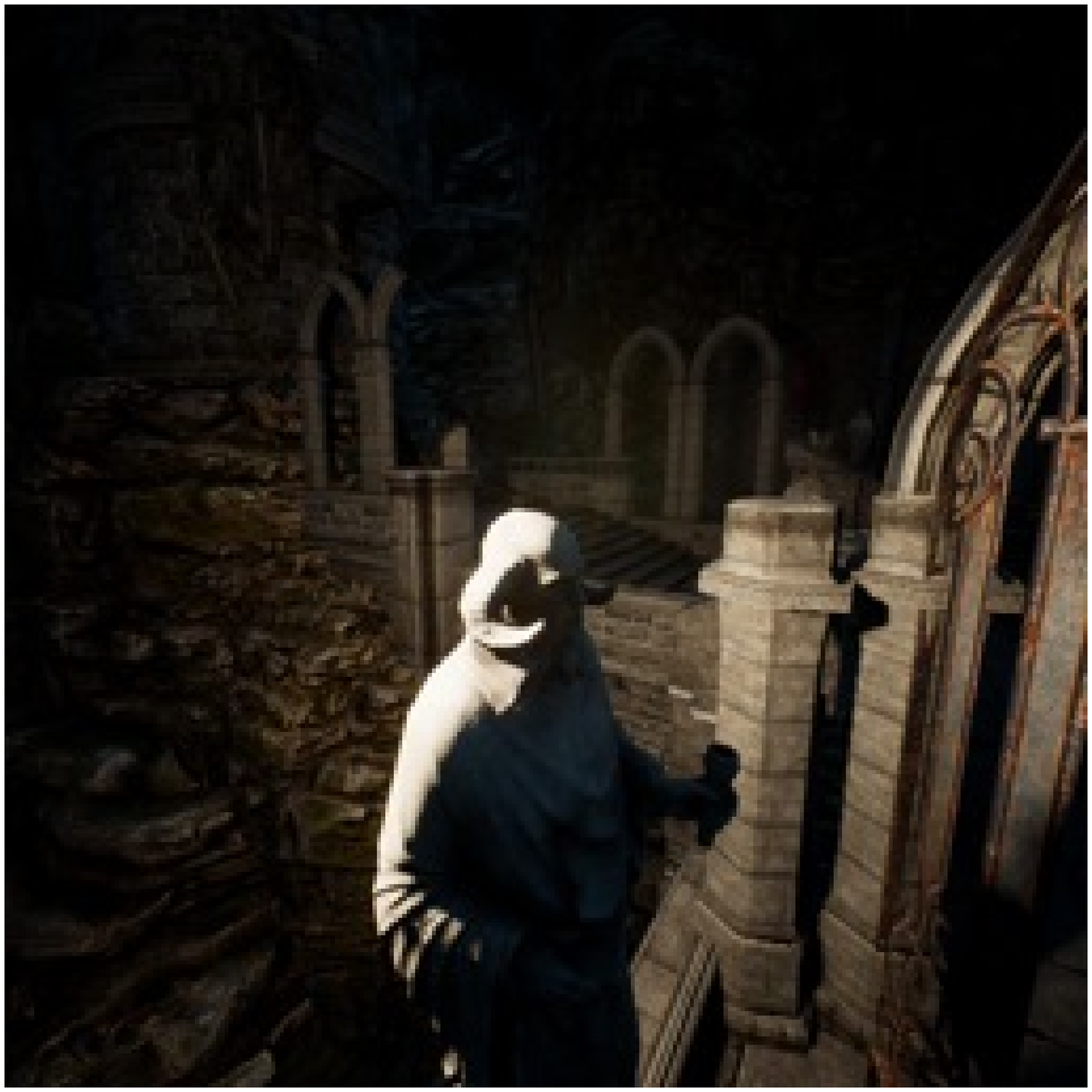} 
\label{fig:subim1}
\end{subfigure}
\begin{subfigure}{0.32\textwidth}
\centering
\includegraphics[width=0.8\linewidth]{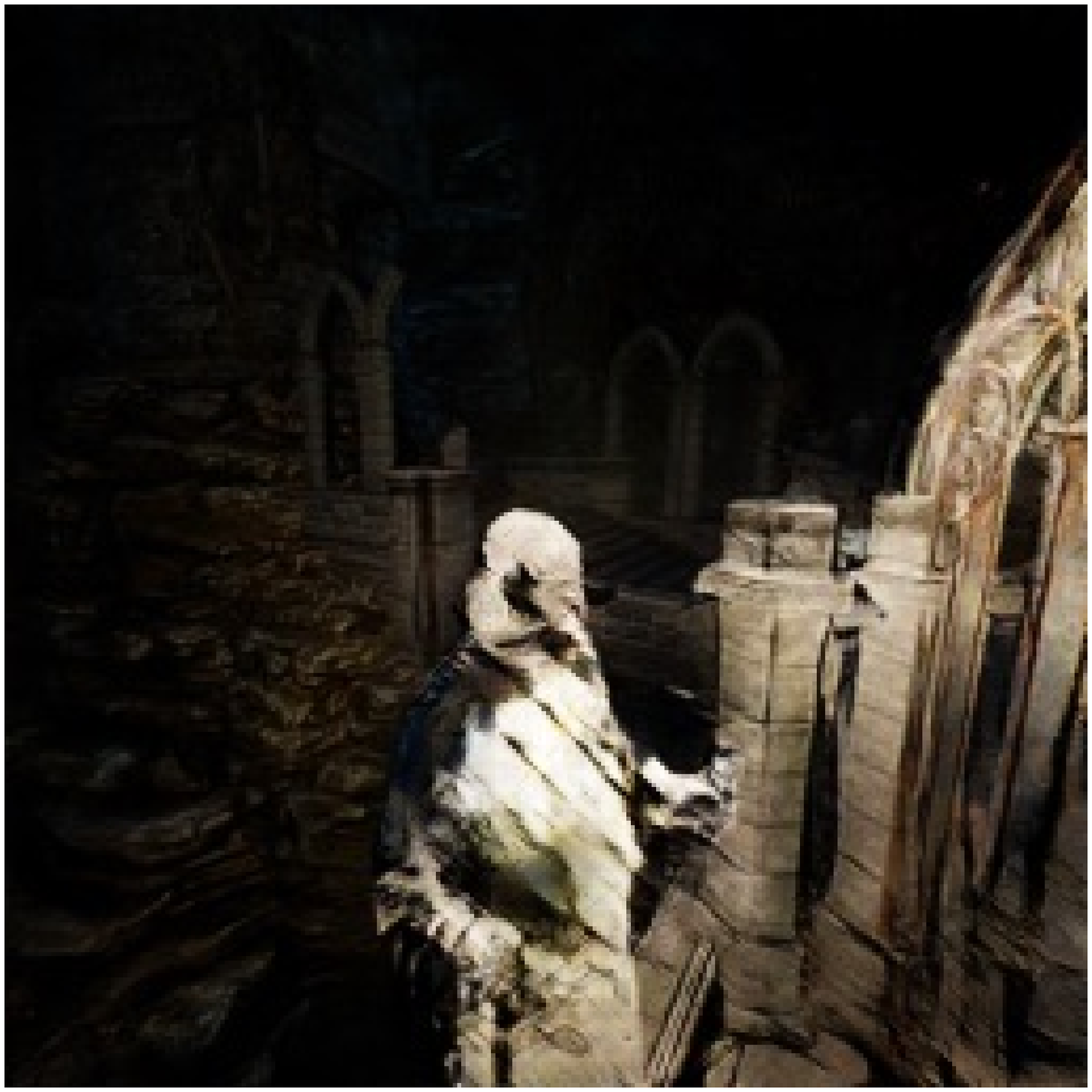} 
\label{fig:subim2}
\end{subfigure}
\begin{subfigure}{0.32\textwidth}
\centering
\includegraphics[width=0.8\linewidth]{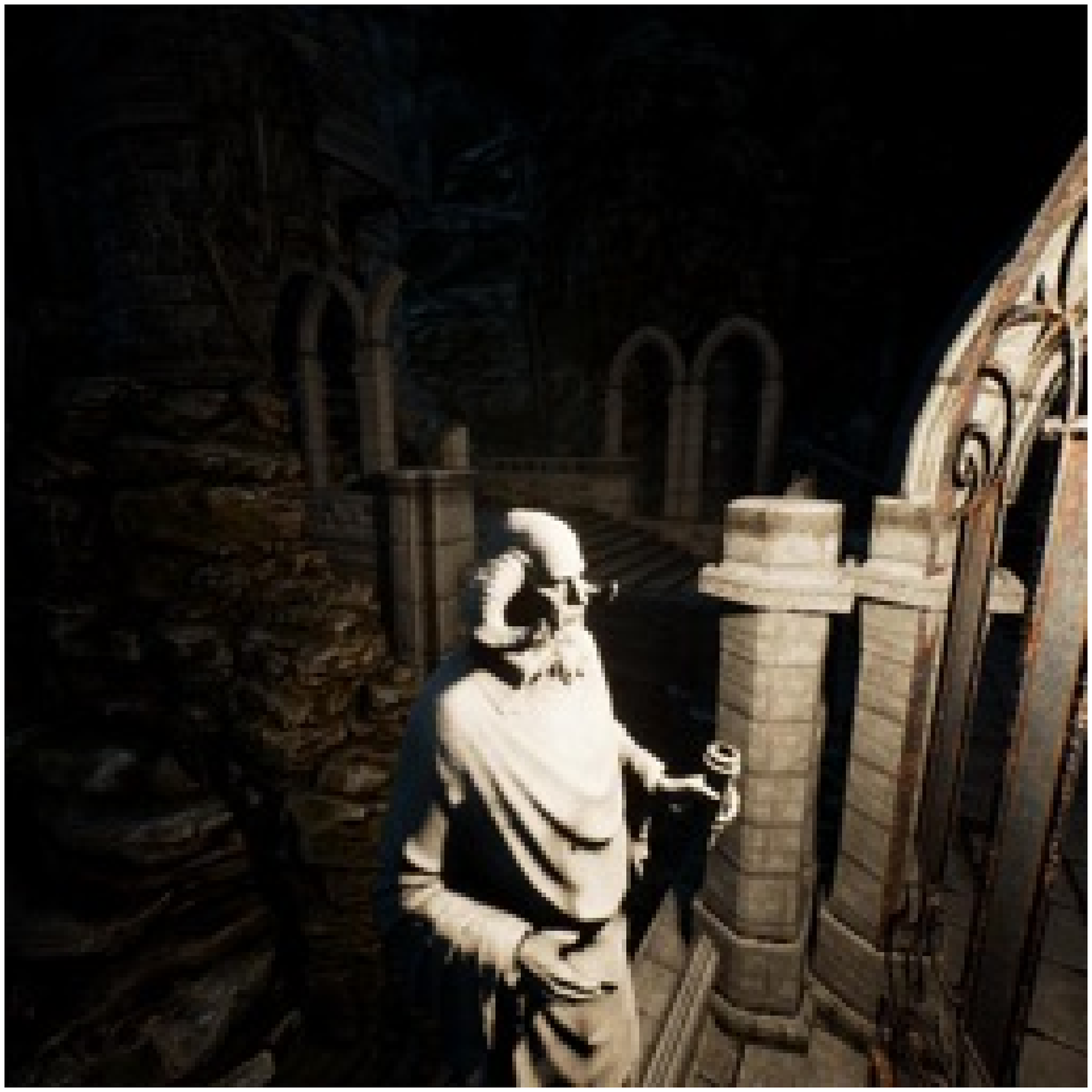} 
\label{fig:subim2}
\end{subfigure}

\begin{subfigure}{0.32\textwidth}
\centering
\includegraphics[width=0.8\linewidth]{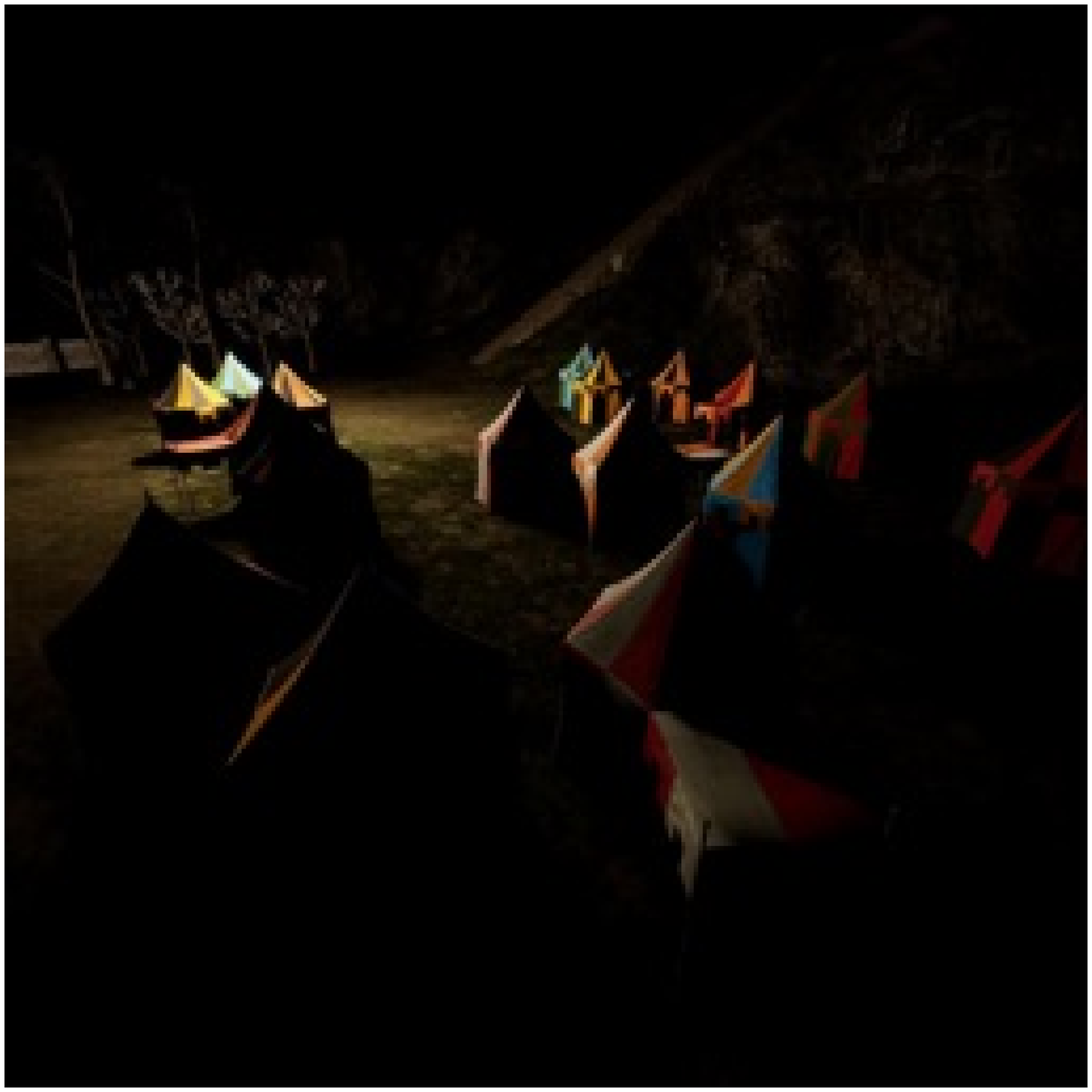} 
\label{fig:subim1}
\end{subfigure}
\begin{subfigure}{0.32\textwidth}
\centering
\includegraphics[width=0.8\linewidth]{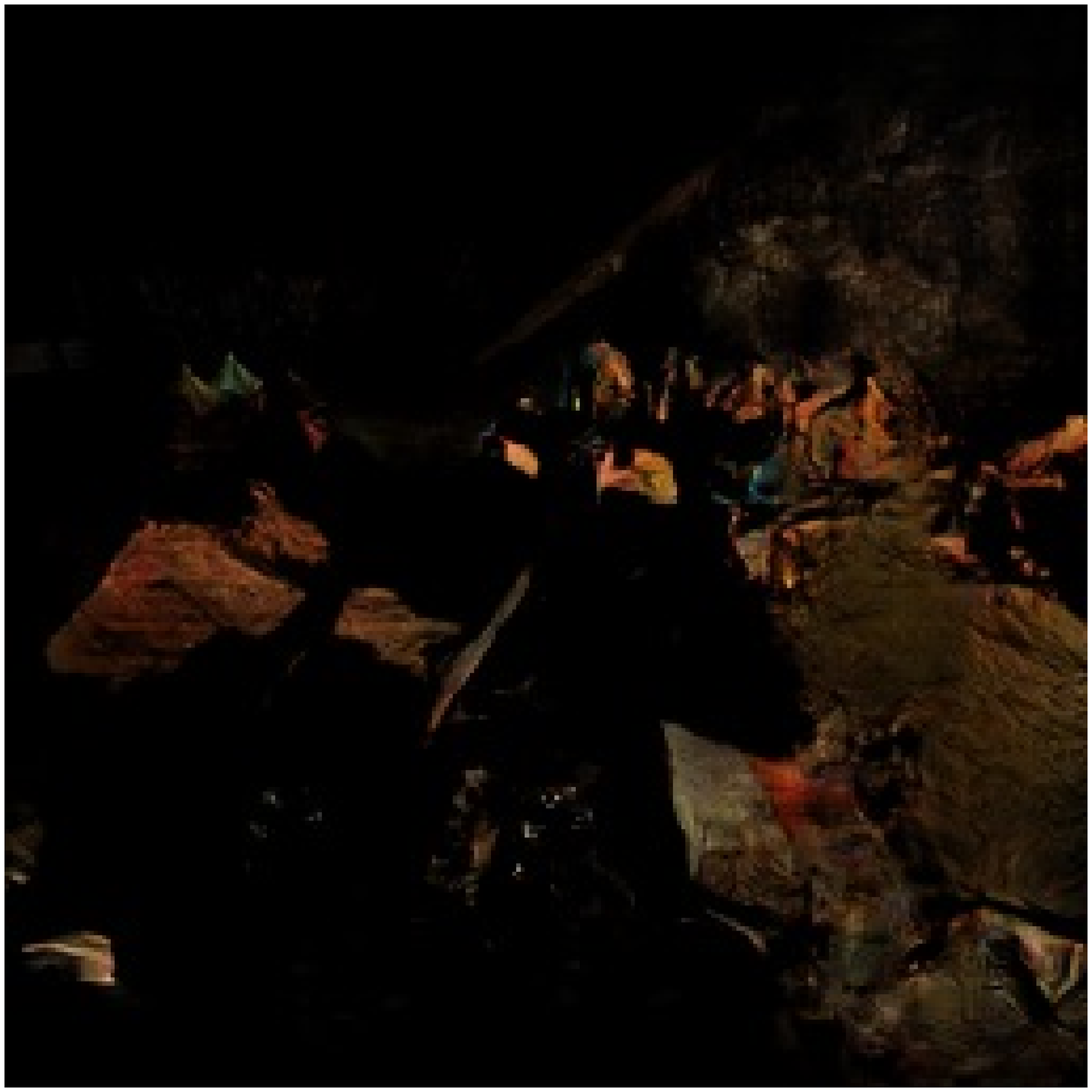} 
\label{fig:subim2}
\end{subfigure}
\begin{subfigure}{0.32\textwidth}
\centering
\includegraphics[width=0.8\linewidth]{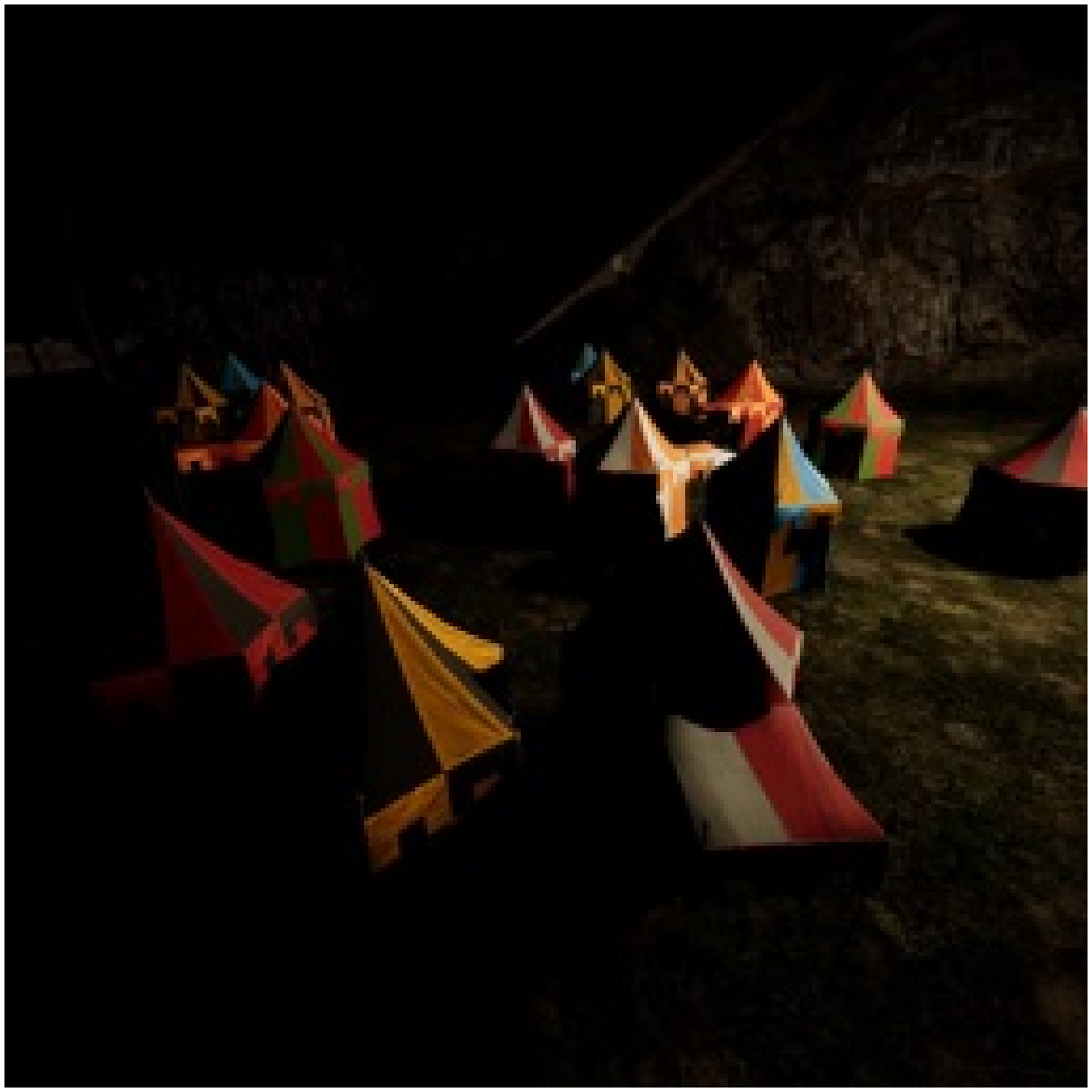} 
\label{fig:subim2}
\end{subfigure}

\begin{subfigure}{0.32\textwidth}
\centering
\includegraphics[width=0.8\linewidth]{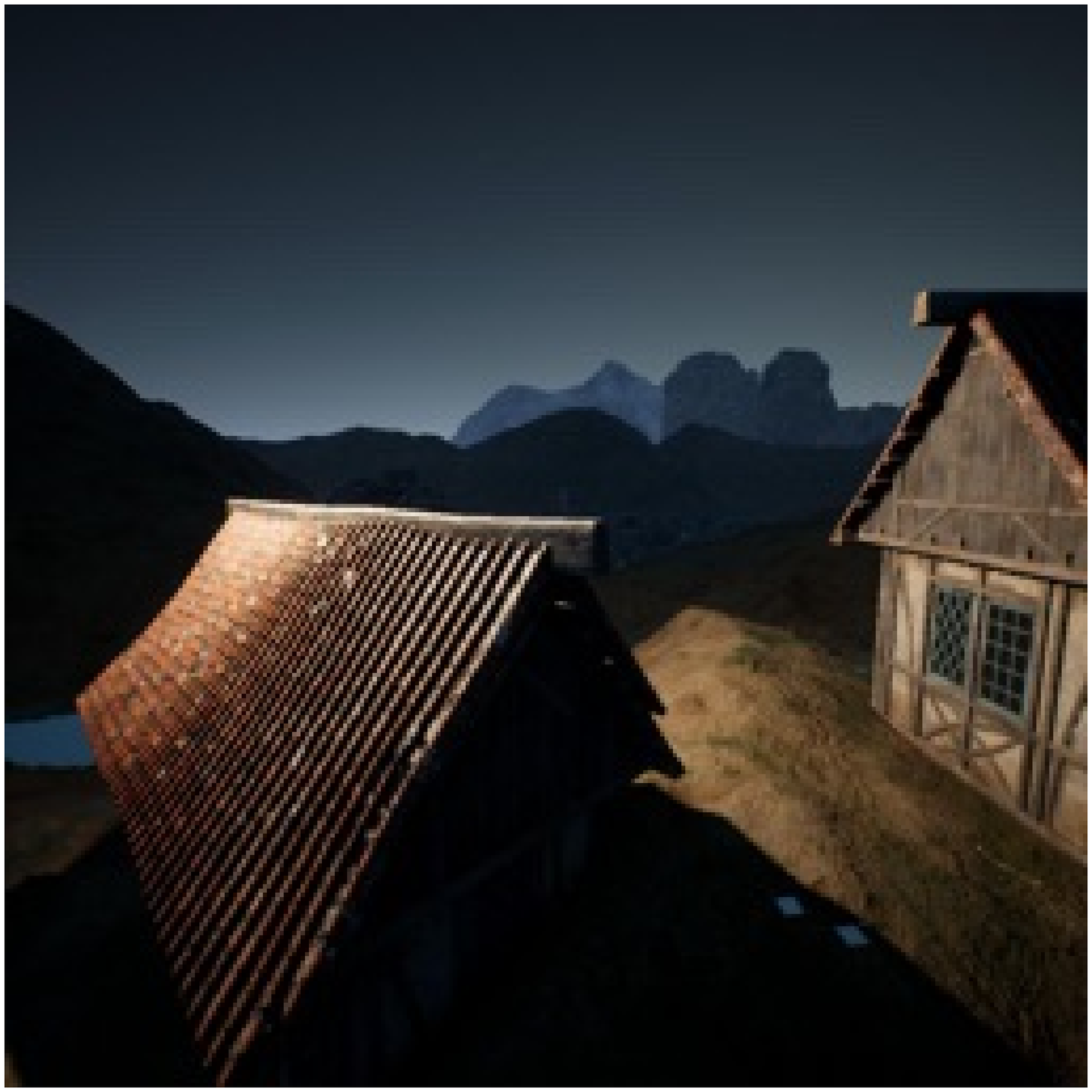} 
\label{fig:subim1}
\caption{Input}
\end{subfigure}
\begin{subfigure}{0.32\textwidth}
\centering
\includegraphics[width=0.8\linewidth]{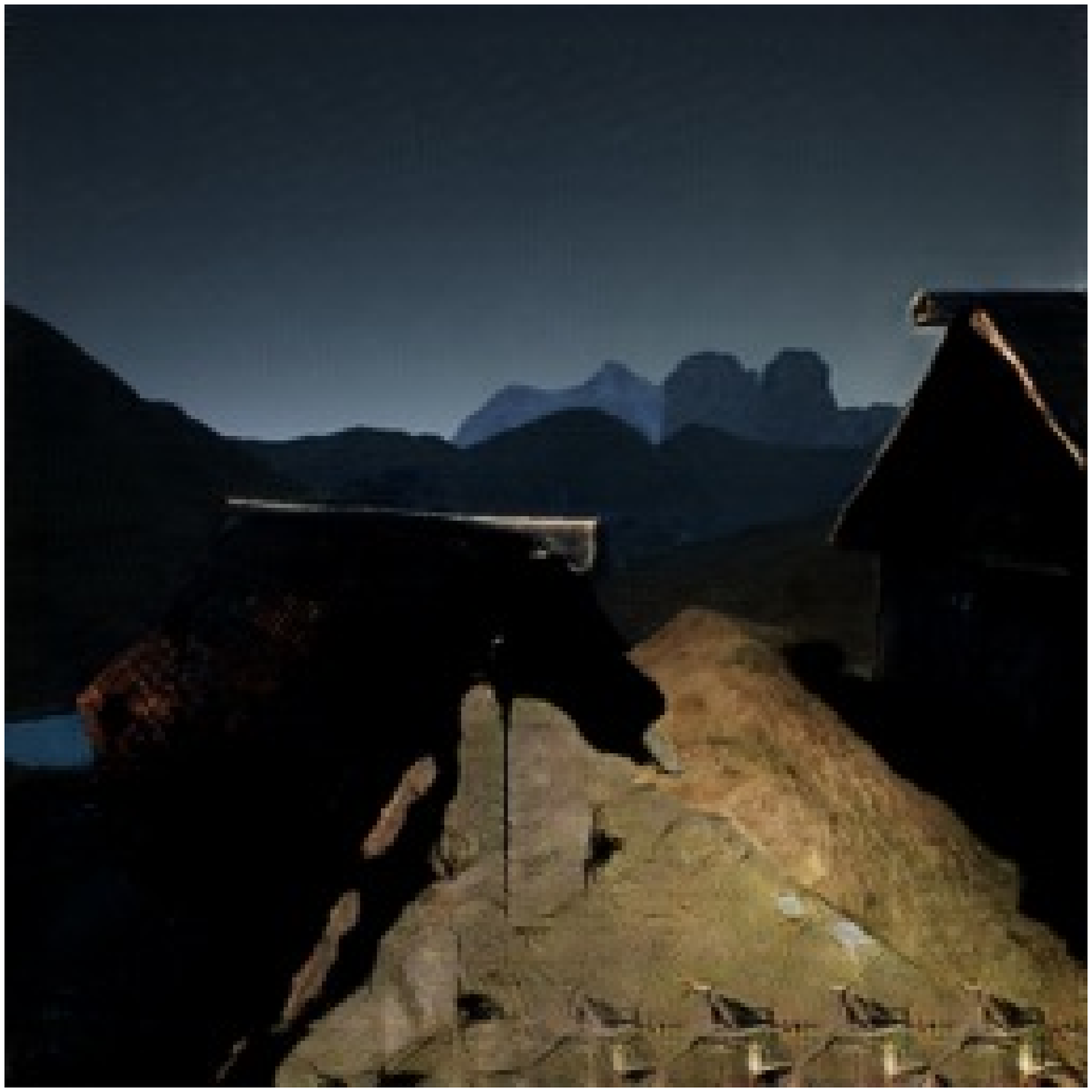} 
\label{fig:subim2}
\caption{Output}
\end{subfigure}
\begin{subfigure}{0.32\textwidth}
\centering
\includegraphics[width=0.8\linewidth]{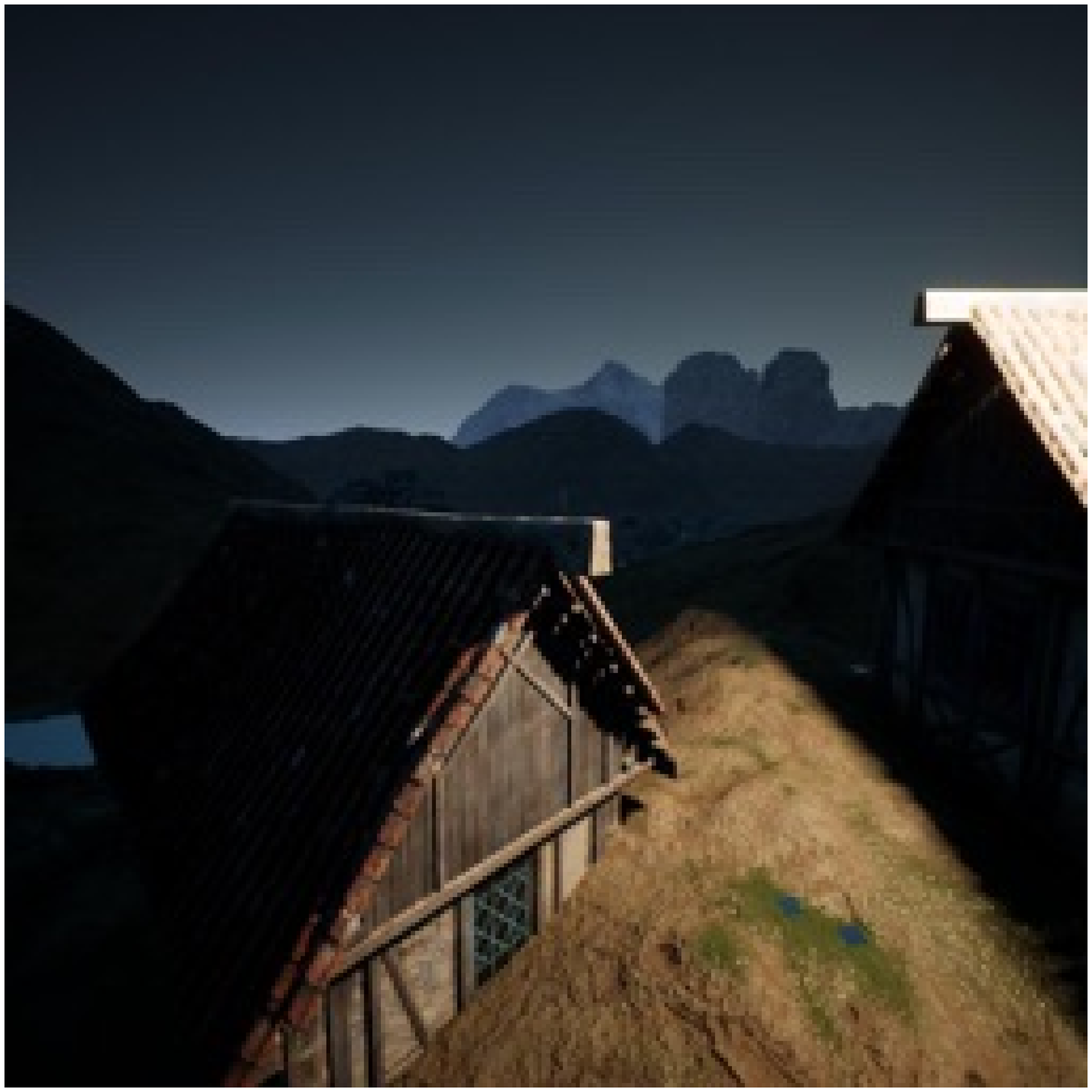} 
\label{fig:subim2}
\caption{Ground Truth}
\end{subfigure}

\caption{Failures, with high distortions and inaccurate shadows}
\label{fig:relight_net2}
\end{figure}

We show examples of these observations from our test data in \autoref{fig:relight_net} and \autoref{fig:relight_net2}. The PSNR (Peak Signal to Noise Ratio) for all the relighting networks is shown in Table \ref{table:psnr}.
%\vspace{2em}
\begin{table}
\centering
\begin{tabular}{| l | c | c | c|}
  \hline
  Target Dir. & PSNR mean (dB) & PSNR std \\
  \hline
  4500K-E & 21.45 &3.63 \\
  4500K-NE & 21.74 &4.75 \\
  4500K-N & 21.55 &4.05 \\
  4500K-NW & 21.36 &4.10 \\
  4500K-W & 20.56 &2.79 \\
  4500K-SW & 21.59 &3.72 \\
  4500K-S & 22.24 &3.71 \\
  4500K-SE & 21.71 &3.48 \\
  \hline
  
\end{tabular}
\vspace{1em}

\caption{PSNR values for each of the target light directions}

\label{table:psnr}
\end{table}

\newpage

\subsection{Evolution of the Relighting Networks}

An interesting observation is the network evolution during training. Each model was trained for about one week, using 1000 epochs. We show below some results obtained at different points in training.

\begin{figure}[h!]
\centering
\begin{subfigure}{0.2\textwidth}
\includegraphics[width=\linewidth]{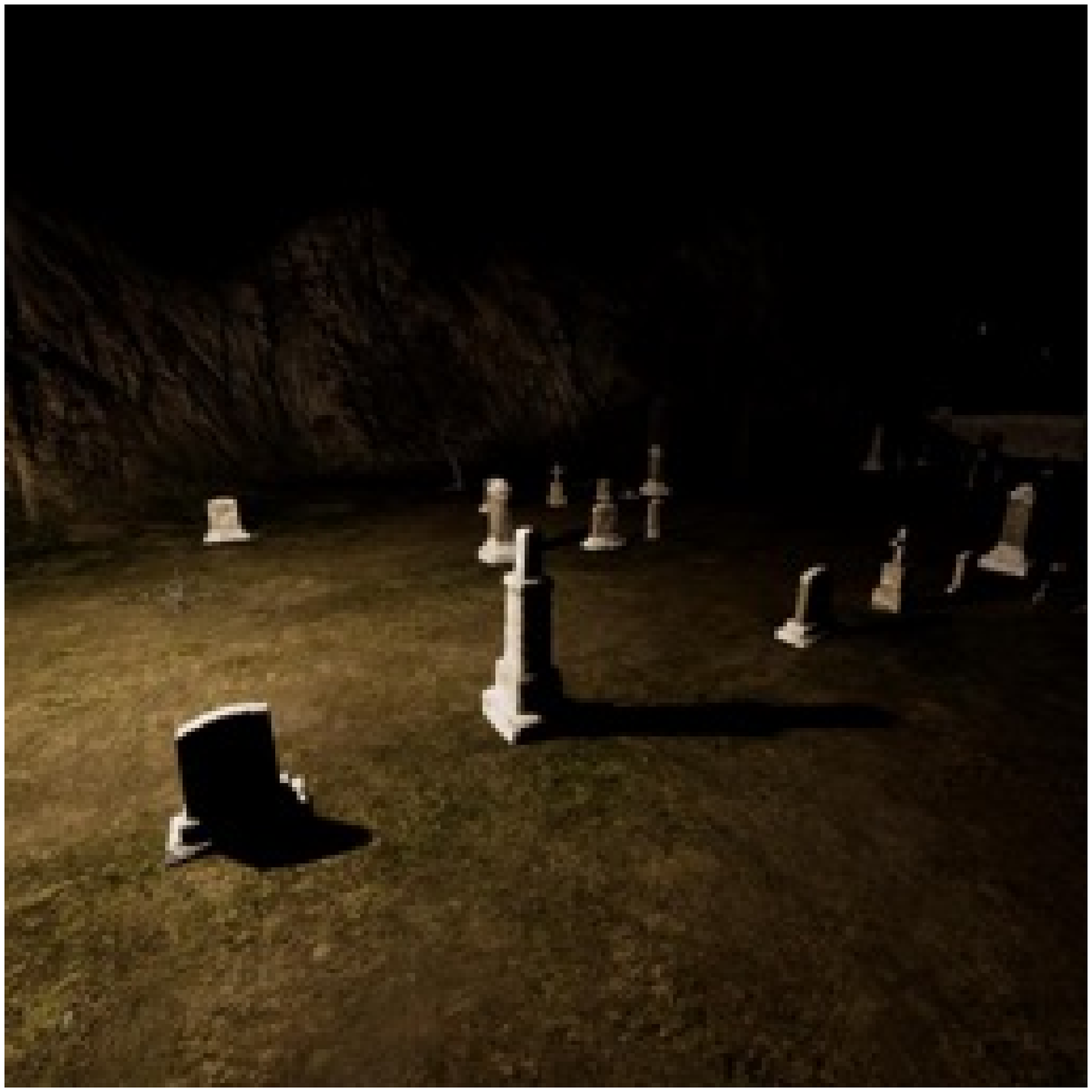} 
\caption{Input}
\end{subfigure}
\begin{subfigure}{0.2\textwidth}
\includegraphics[width=\linewidth]{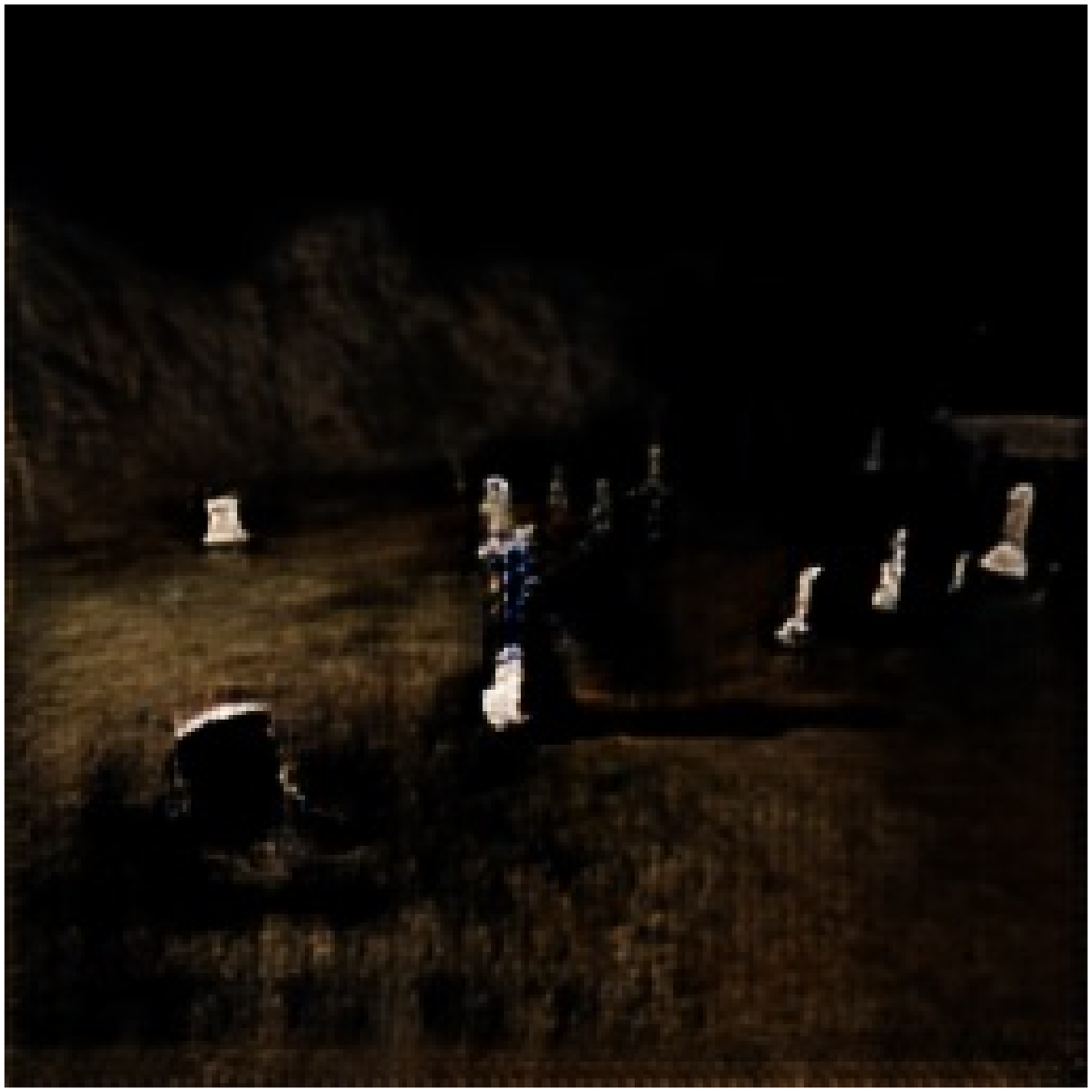} 
\caption{Epoch 10}
\end{subfigure}
\begin{subfigure}{0.2\textwidth}
\includegraphics[width=\linewidth]{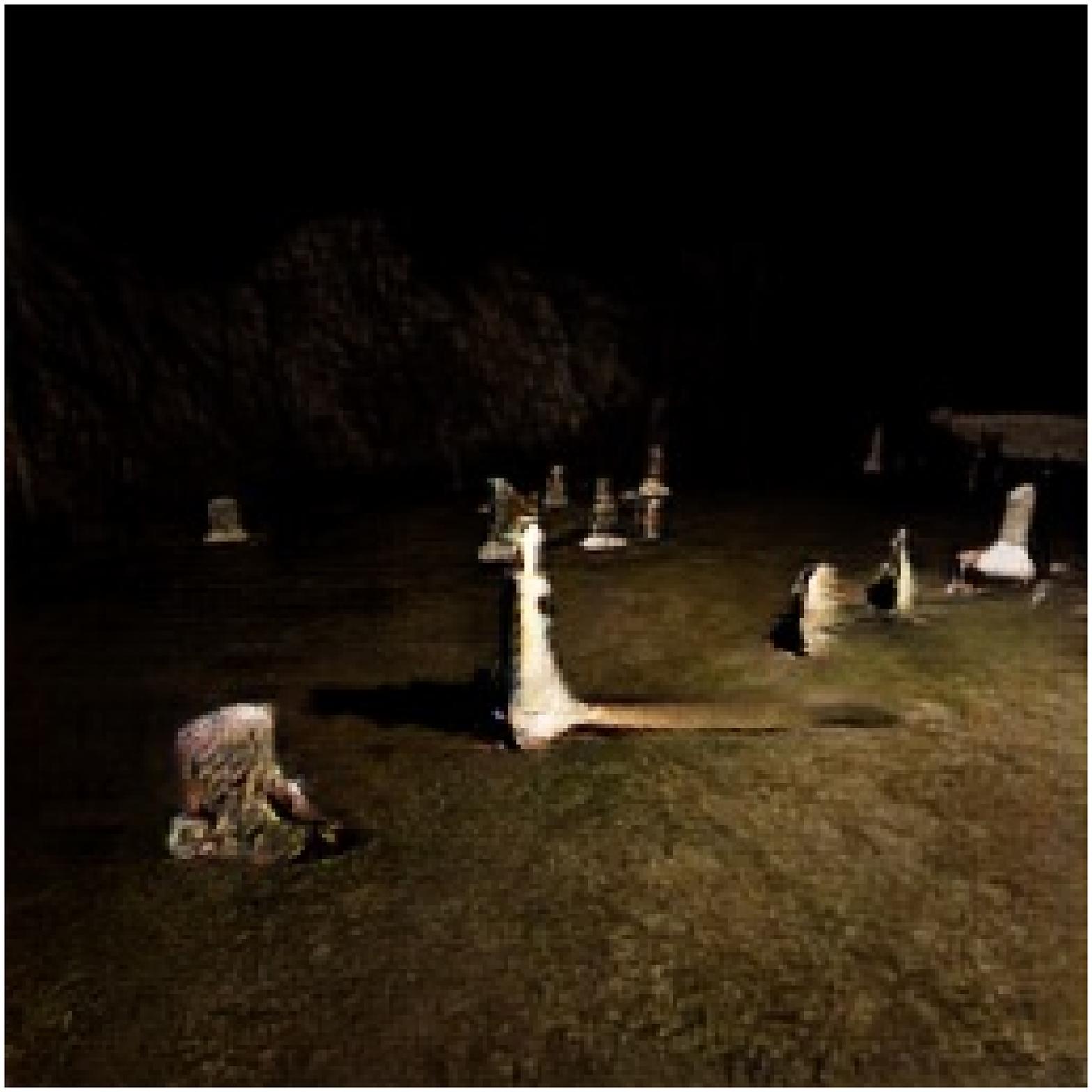} 
\caption{Epoch 100}
\end{subfigure}
\begin{subfigure}{0.2\textwidth}
\includegraphics[width=\linewidth]{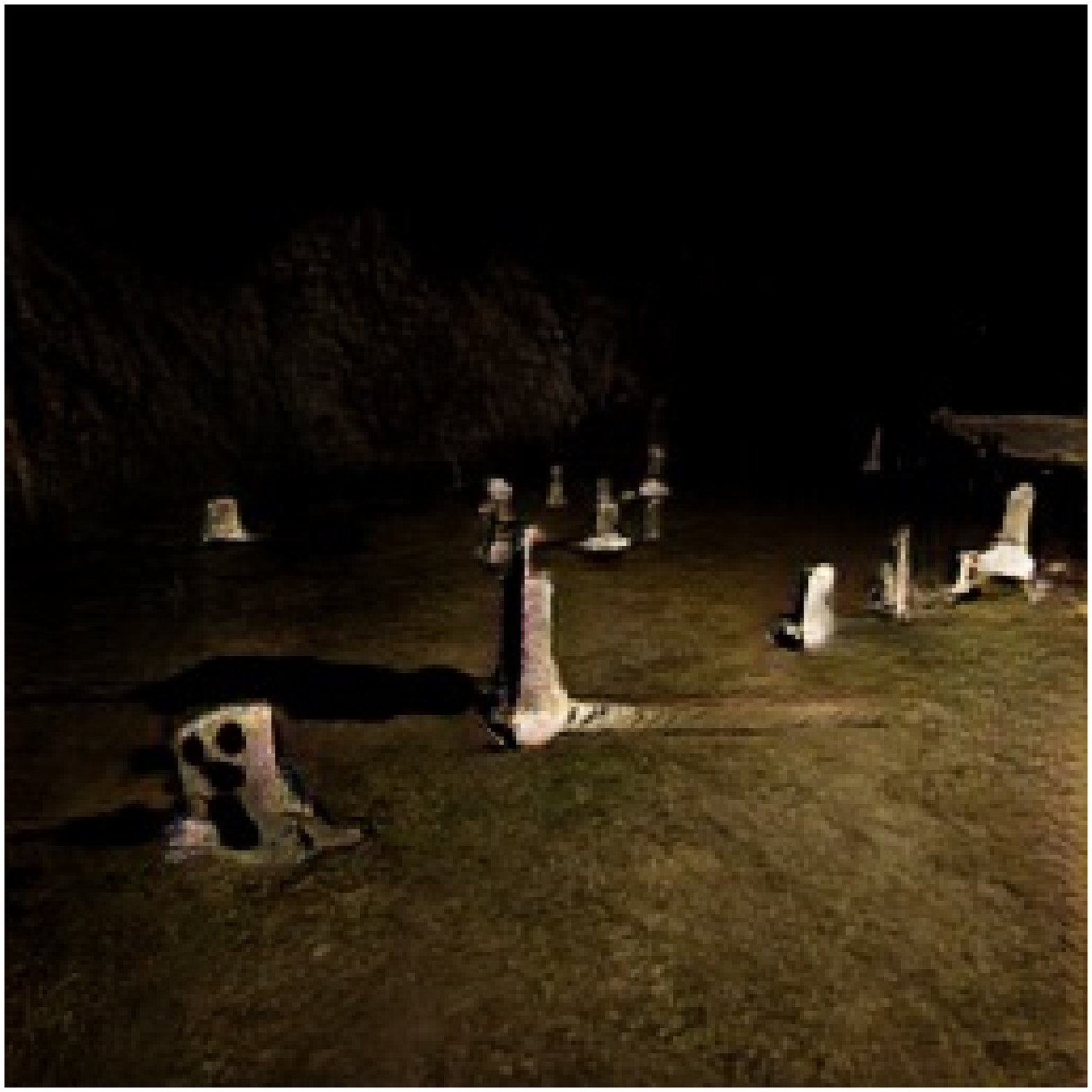} 
\caption{Epoch 250}
\end{subfigure}

\begin{subfigure}{0.2\textwidth}
\includegraphics[width=\linewidth]{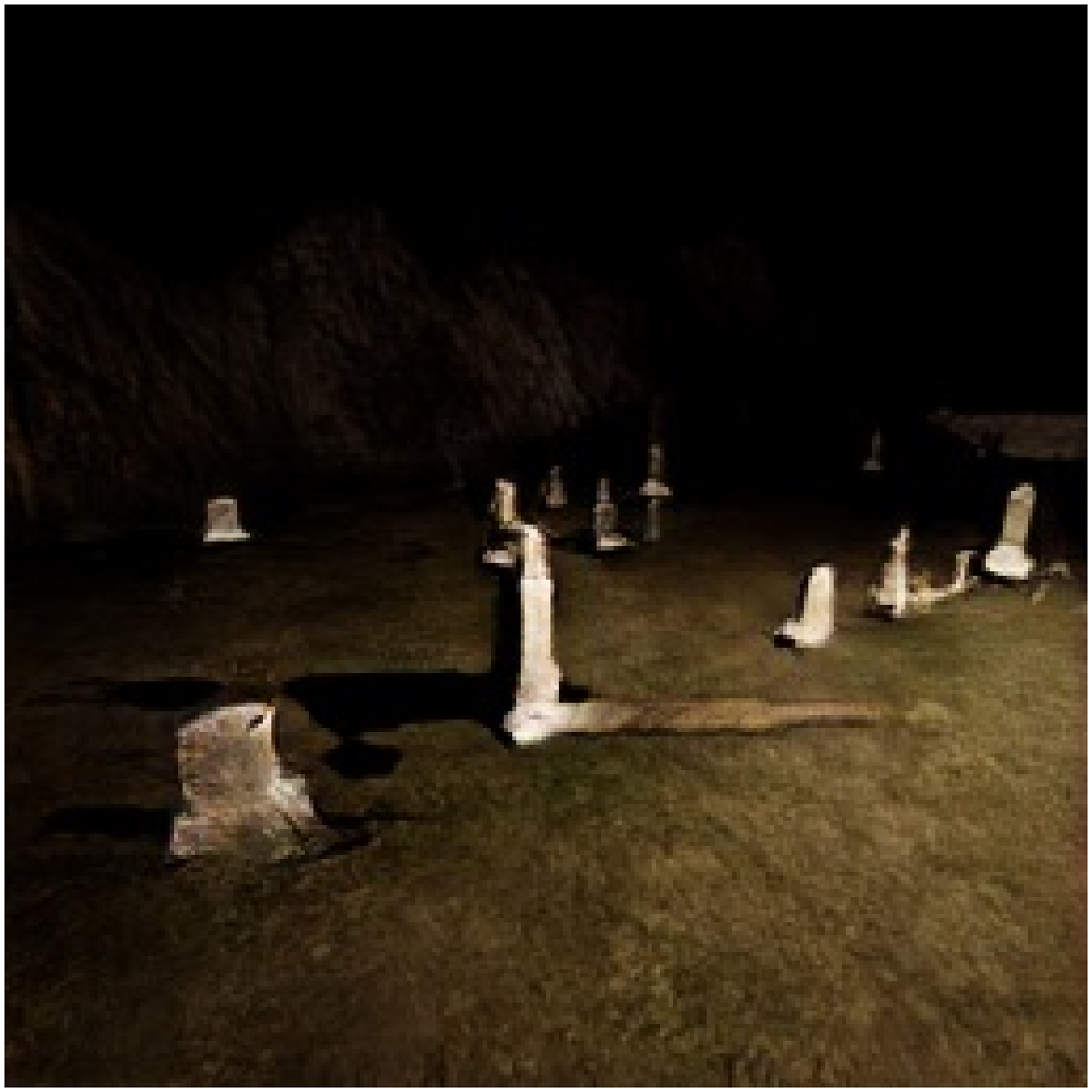} 
\caption{Epoch 500}
\end{subfigure}
\begin{subfigure}{0.2\textwidth}
\includegraphics[width=\linewidth]{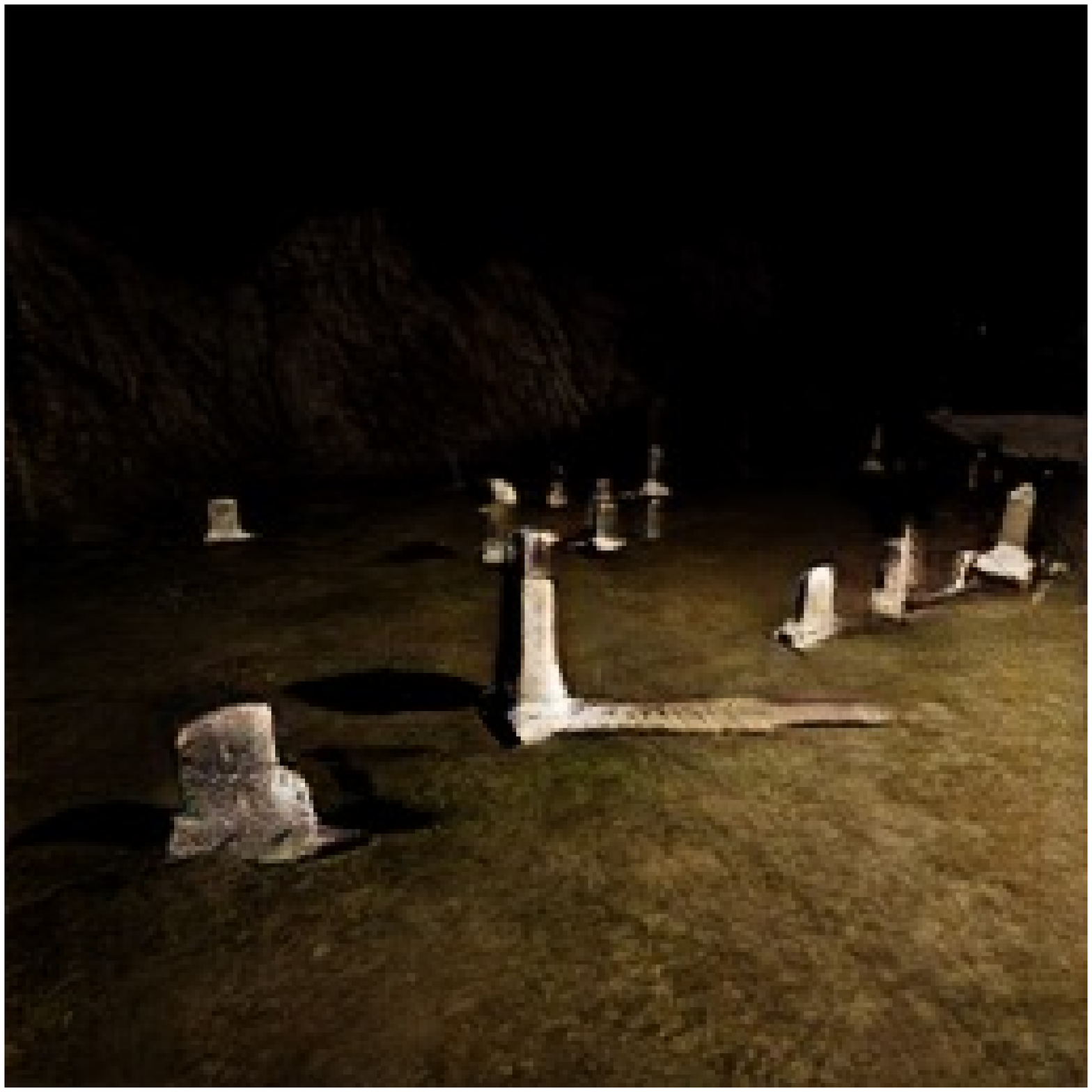} 
\caption{Epoch 750}
\end{subfigure}
\begin{subfigure}{0.2\textwidth}
\includegraphics[width=\linewidth]{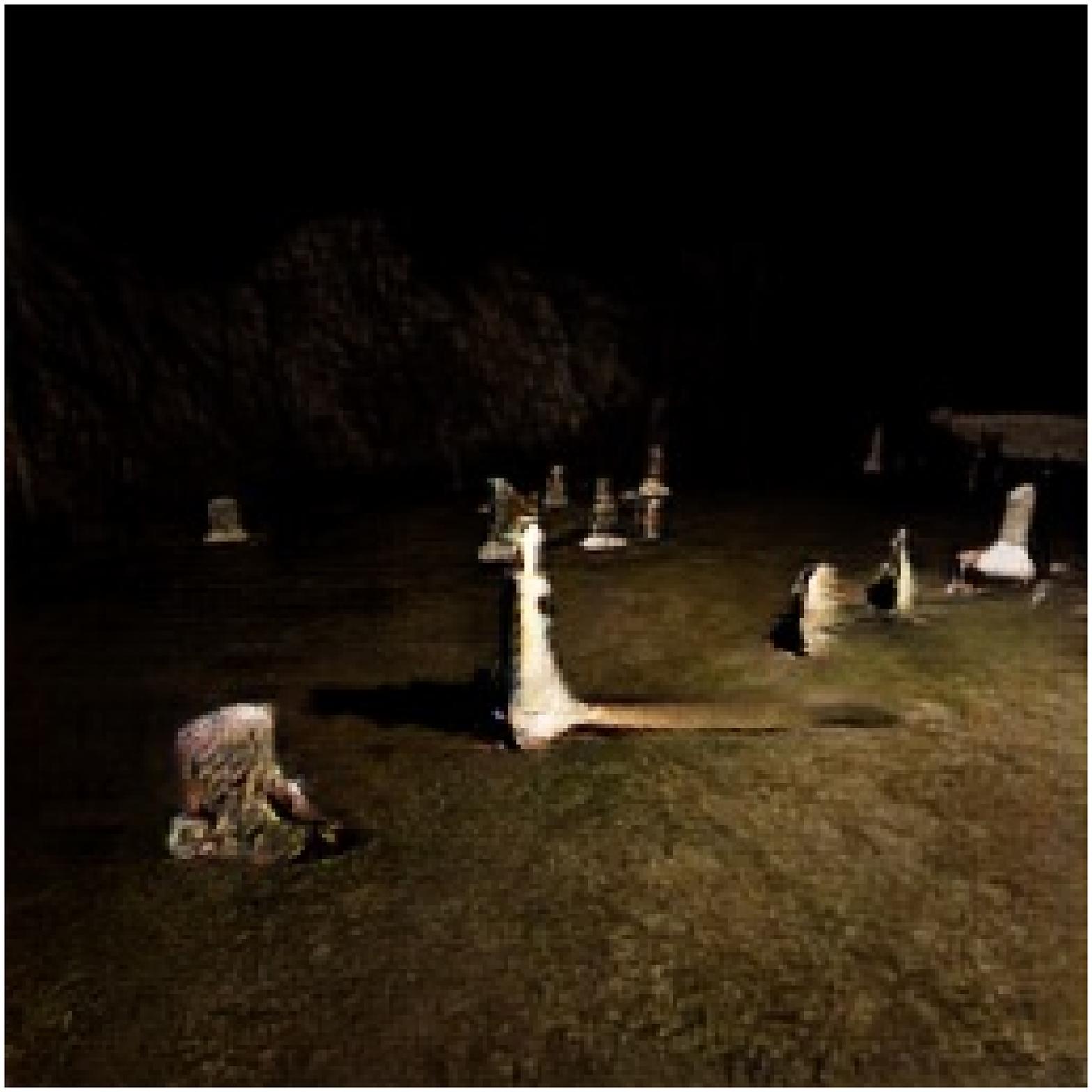} 
\caption{Epoch 1000}
\end{subfigure}
\begin{subfigure}{0.2\textwidth}
\includegraphics[width=\linewidth]{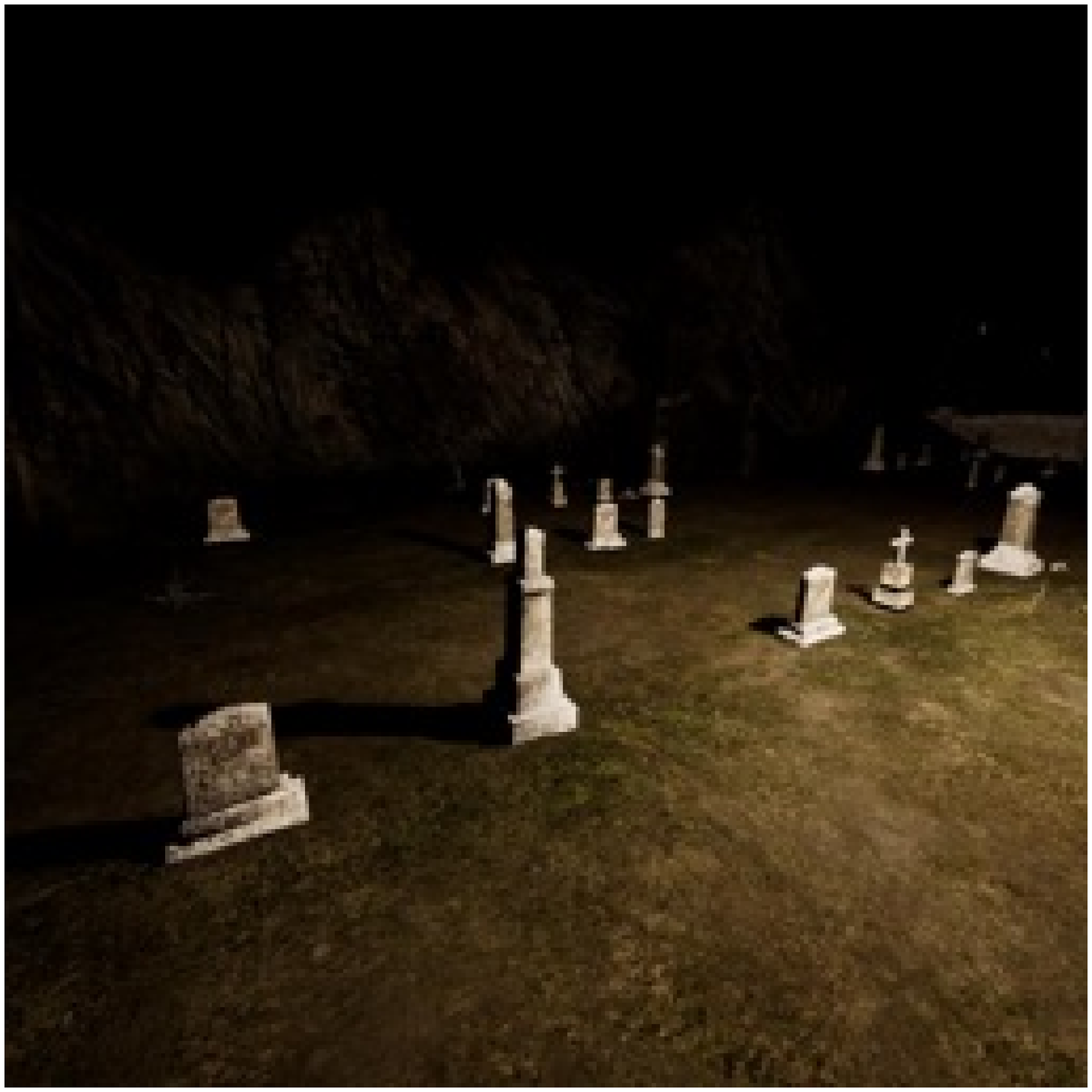} 
\caption{GT}
\end{subfigure}

\caption{The evolution of the relighting network on test image \textbf{scene\_grave\_43\_W\_E\_11}, with lighting change from West to East}
\label{fig:epochs1}
\end{figure}

\begin{figure}[h!]
\centering
\begin{subfigure}{0.2\textwidth}
\includegraphics[width=\linewidth]{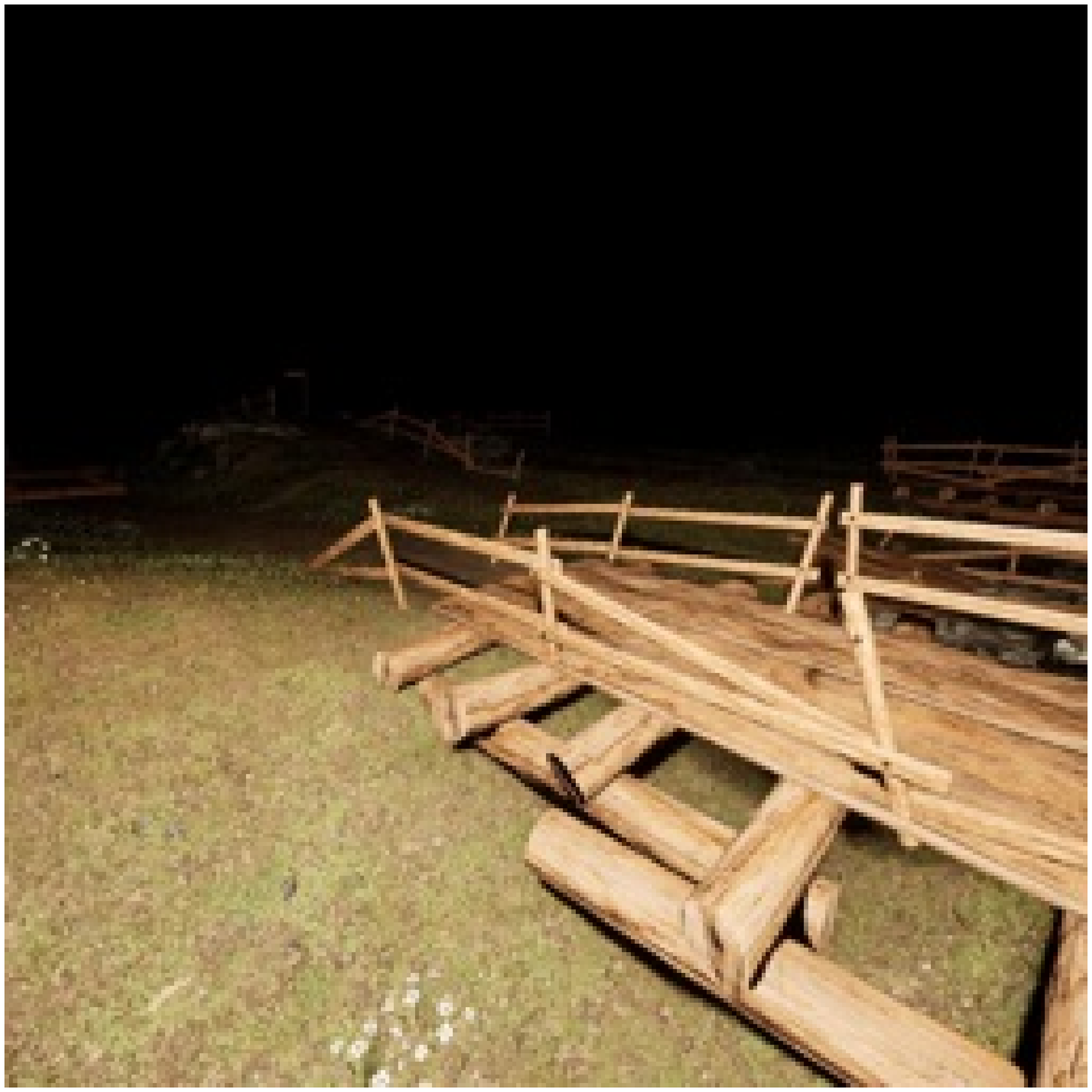} 
\caption{Input}
\end{subfigure}
\begin{subfigure}{0.2\textwidth}
\includegraphics[width=\linewidth]{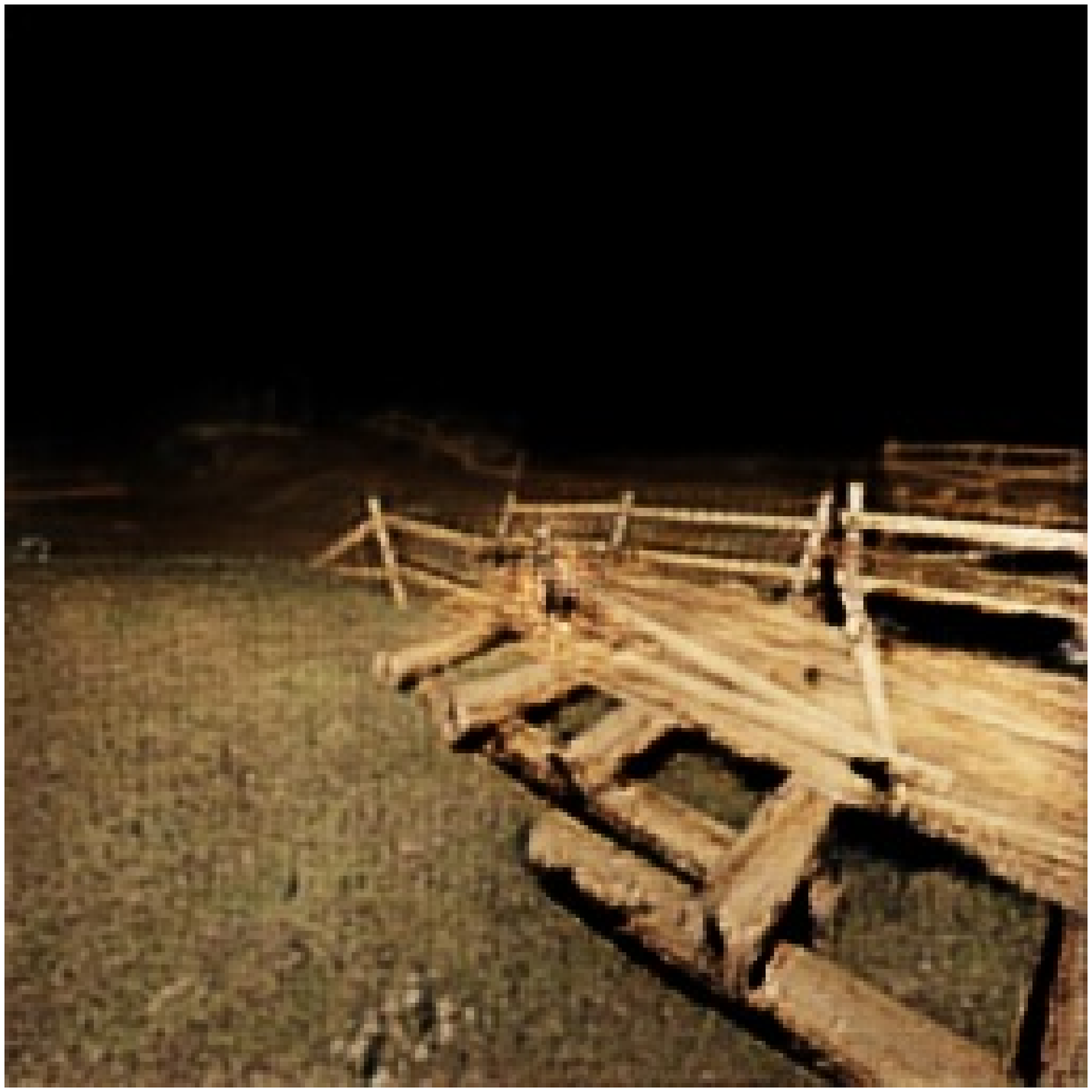} 
\caption{Epoch 10}
\end{subfigure}
\begin{subfigure}{0.2\textwidth}
\includegraphics[width=\linewidth]{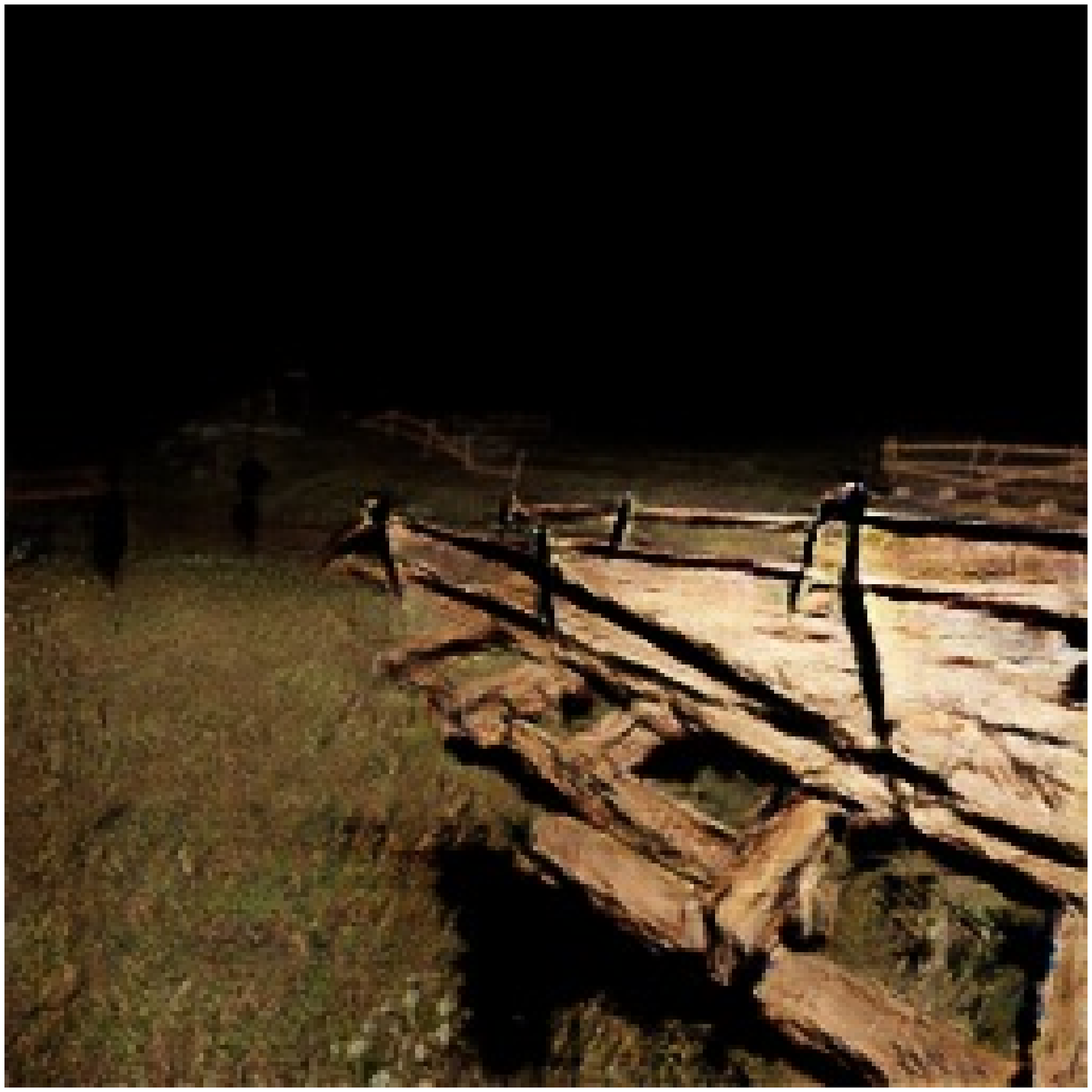} 
\caption{Epoch 100}
\end{subfigure}
\begin{subfigure}{0.2\textwidth}
\includegraphics[width=\linewidth]{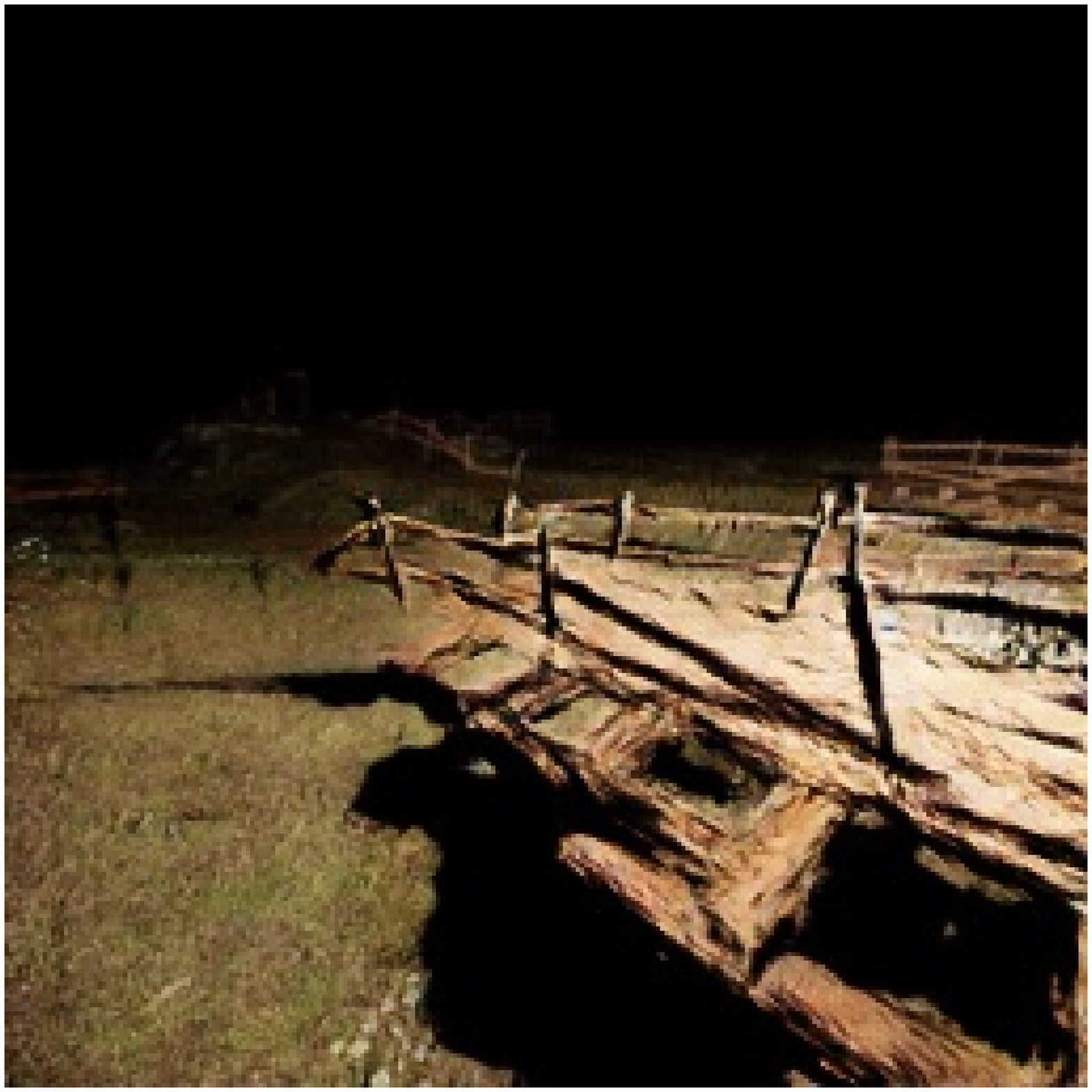} 
\caption{Epoch 250}
\end{subfigure}

\begin{subfigure}{0.2\textwidth}
\includegraphics[width=\linewidth]{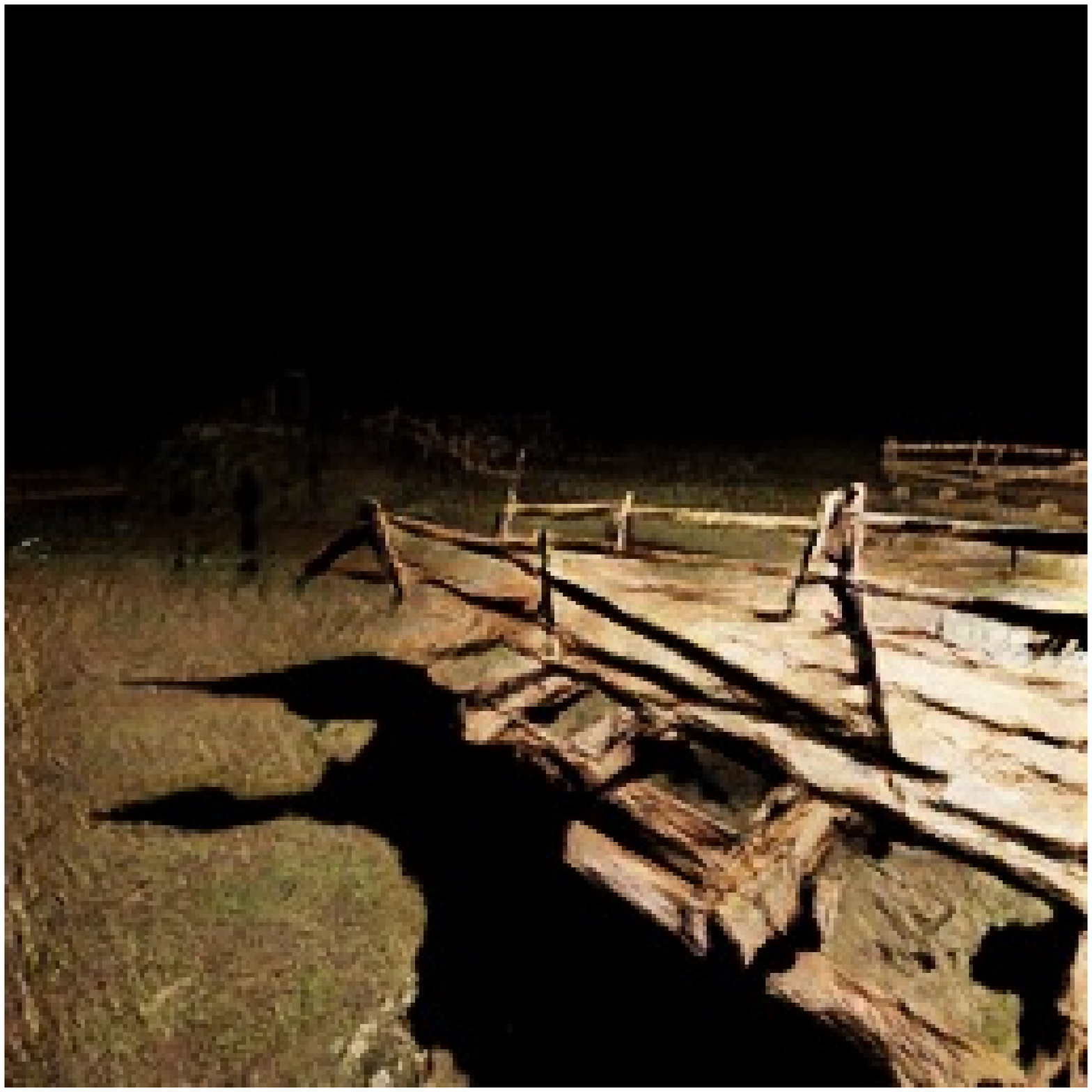} 
\caption{Epoch 500}
\end{subfigure}
\begin{subfigure}{0.2\textwidth}
\includegraphics[width=\linewidth]{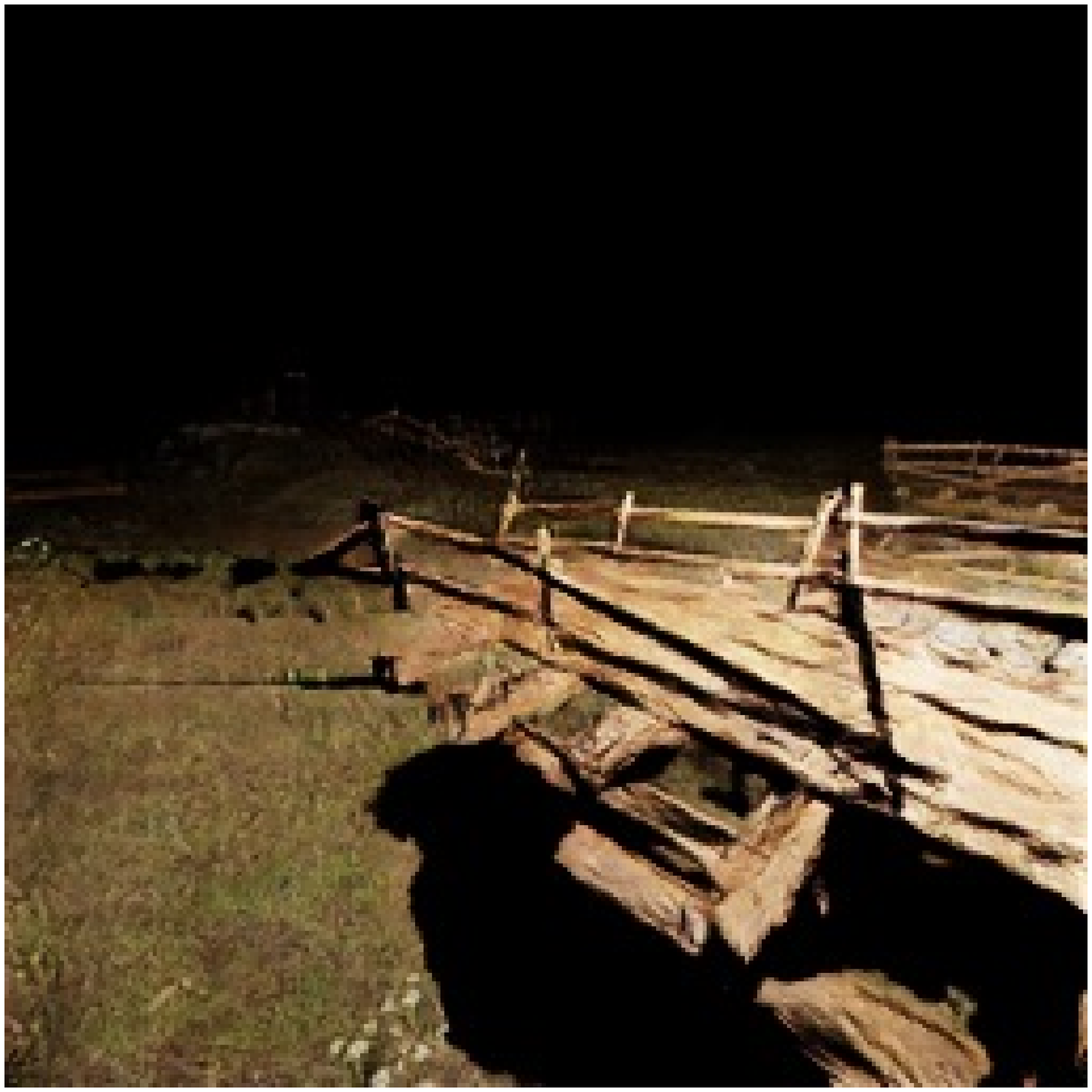} 
\caption{Epoch 750}
\end{subfigure}
\begin{subfigure}{0.2\textwidth}
\includegraphics[width=\linewidth]{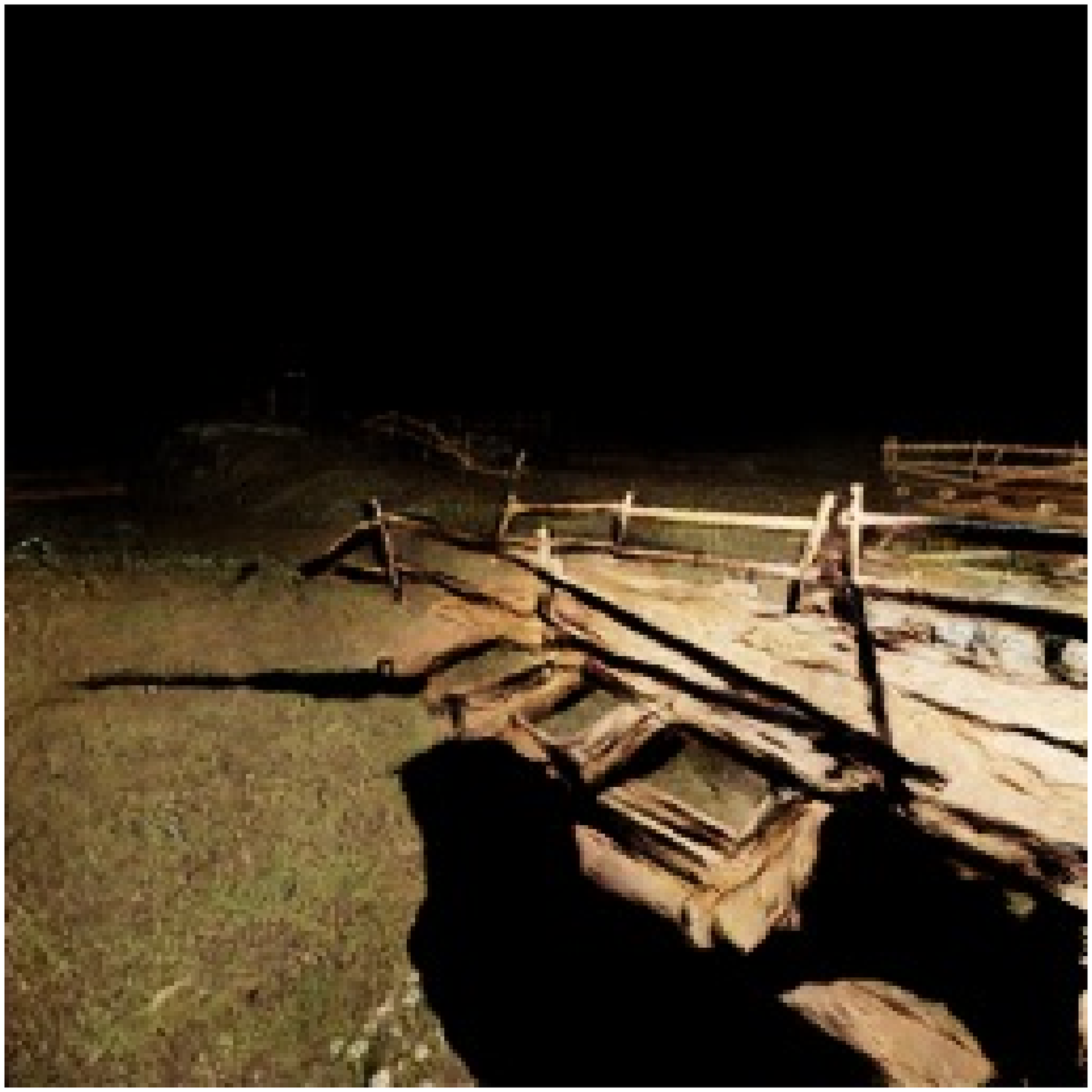} 
\caption{Epoch 1000}
\end{subfigure}
\begin{subfigure}{0.2\textwidth}
\includegraphics[width=\linewidth]{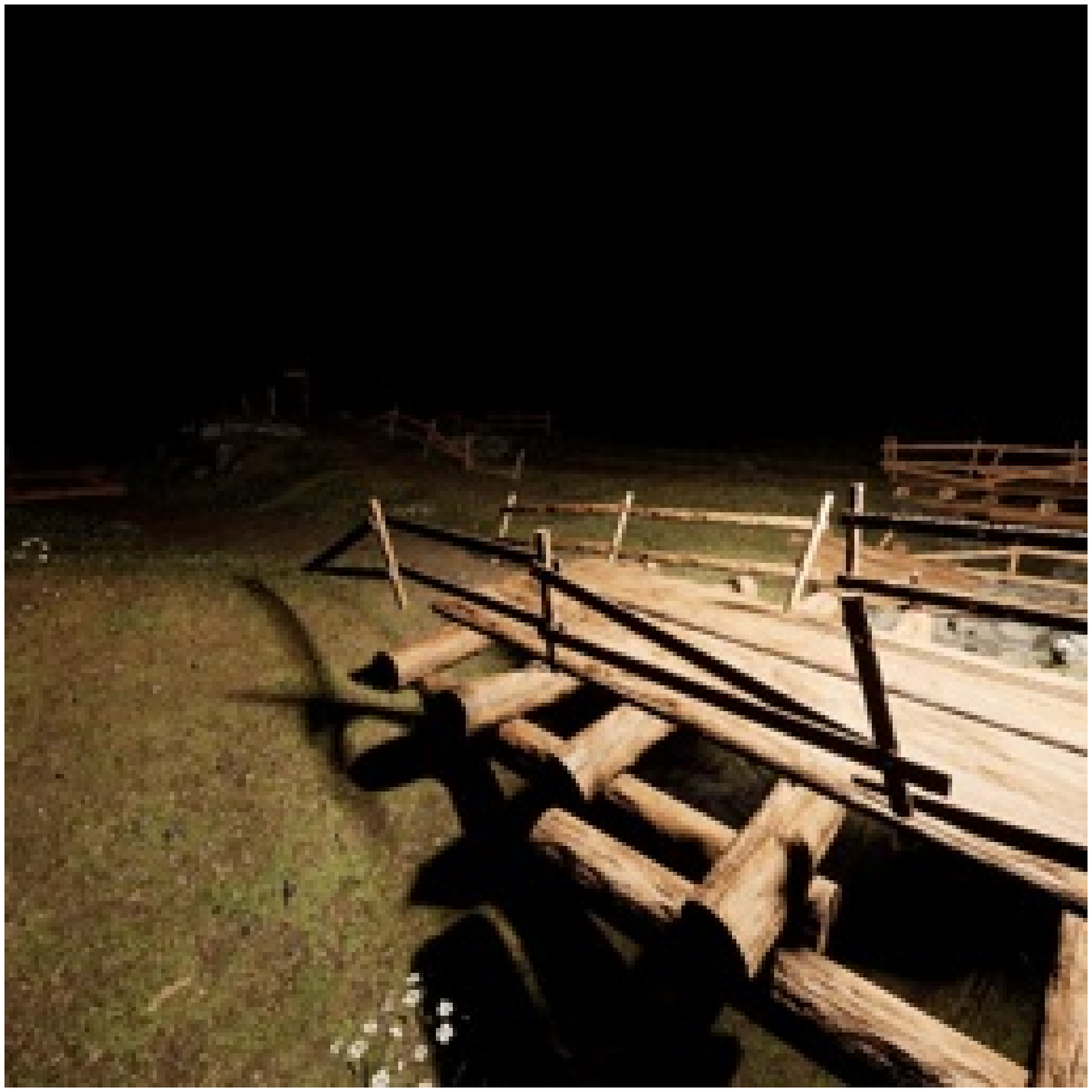} 
\caption{GT}
\end{subfigure}

\caption{The evolution of the relighting network on test image \textbf{scene\_road\_33\_4500\_S\_E\_16}, with lighting change from South to East}
\label{fig:epochs2}
\end{figure}

\subsection{Light Direction Classifier}

The light direction classifier has been trained as an eight-class classifier, each class representing one of the eight possible light directions. It has been trained on an 80\%-20\% train-test split with an Adam optimizer and variable learning rate. The accuracy is 65\%, with 97\% of the images being either correctly classified or assigned a neighbouring light direction. This means that 65\% of the test images are classified correctly within a forty-five-degree range and 97\% are classified correctly within a ninety-degree range.

\section{Conclusions and Future Work}

Image relighting in a 2D setting is a very difficult problem, due to the limited information available. We have shown that a simple image-to-image translation can solve this problem up to a certain extent. We believe that having a more realistic dataset, with more indirect lighting such that the shadows contain more information would greatly improve the results of this approach. Moreover, adding some explicit 3D information would most likely lead to significant improvements. We shortly experimented in this direction, following the idea of \cite{3dScene} and replacing the geometric prior with a monocular depth estimation. While depth estimation has good results from the camera viewpoint, the model does not capture enough detail from the side views, suffering from "soap film" artifacts, which leads to extremely bad shadows. There is however more advanced research into this area, with works estimating the full 3D shape from a single view and not only the depth, such as \cite{DBLP:journals/corr/abs-1803-11493}.

\clearpage

\bibliographystyle{splncs}
\bibliography{egbib}
\end{document}